\documentclass[10pt,journal,compsoc]{IEEEtran}
\usepackage{graphicx}
\usepackage{amsmath,amssymb} % define this before the line numbering.

\usepackage[pagebackref=true,breaklinks=true,letterpaper=true,colorlinks,bookmarks=false]{hyperref}

\usepackage{times}
\usepackage{helvet}
\usepackage{courier}
\usepackage{graphicx}
\usepackage{bm}
\usepackage{amsmath,amssymb} % define this before the line numbering.
\usepackage{color}
\usepackage{bbm}
\usepackage{epstopdf}
\usepackage{caption}
\usepackage{subcaption}
\usepackage{enumitem}
\usepackage{calc}
\usepackage{multirow}
\usepackage{xspace}
\usepackage{booktabs}
\usepackage{mathrsfs}
\usepackage{array}
\usepackage{float}
\usepackage[font=small,skip=4pt]{caption}
\usepackage{graphicx}
\usepackage{hyperref}
\usepackage{url}
\usepackage{etoolbox}
\apptocmd{\sloppy}{\hbadness 10000\relax}{}{}

\usepackage{threeparttable}

\newcommand{\figref}[1]{Fig\onedot~\ref{#1}}
\newcommand{\equref}[1]{Eq\onedot~\eqref{#1}}
\newcommand{\secref}[1]{Sec\onedot~\ref{#1}}
\newcommand{\tabref}[1]{Tab\onedot~\ref{#1}}

\newcommand{\ve}[1]{{\mathbf #1}} % for displaying a vector or matrix
\newcommand{\hua}[1]{{\mathcal #1}}

\newcommand{\by}[2]{\ensuremath{#1 \! \times \! #2}}
\newcommand{\thickhline}{%
    \noalign {\ifnum 0=`}\fi \hrule height 1pt
    \futurelet \reserved@a \@xhline
}

\newcommand{\peng}[1]{\textcolor{black}{#1}}
\DeclareRobustCommand\onedot{\futurelet\@let@token\@onedot}
\def\onedot{\ifx\@let@token.\else.\null\fi\xspace}
\def\eg{\emph{e.g., }}

\def\ie{\emph{i.e., }}

\def\etc{\emph{etc}\onedot}

\def\wrt{w.r.t\onedot}

\def\etal{\emph{et al.}}

\newcommand*\rot[1]{\rotatebox{45}{#1}}
\newcommand*\rotf[1]{\rotatebox{35}{#1}}

\makeatletter
\newcommand\footnoteref[1]{\protected@xdef\@thefnmark{\ref{#1}}\@footnotemark}
\makeatother

\ifCLASSOPTIONcompsoc
  \usepackage[nocompress]{cite}
\else
  \usepackage{cite}
\fi
\ifCLASSINFOpdf
\else
\fi
\graphicspath{./fig/}

\begin{document}
\title{The ApolloScape Open Dataset for Autonomous Driving and its Application}


% \author{Xinyu Huang}
% \author{Xinjing Cheng}
% \author{Qichuan Geng}
% \author{Binbin Cao}
% \author{\authorcr Dingfu Zhou}
% \author{Peng Wang}
% \author{Yuanqing Lin}
% \author{Ruigang Yang}

\author{~Xinyu Huang*,
        ~Peng~Wang*,
        ~Xinjing Cheng,
        ~Dingfu Zhou,
       % ~Xibin Song,
        ~Qichuan Geng,
        %~Binbin Cao,
        %~Yuanqing Lin,~\IEEEmembership{Fellow,~IEEE}
        ~Ruigang Yang%,~\IEEEmembership{Fellow,~IEEE}
\IEEEcompsocitemizethanks{
\IEEEcompsocthanksitem X. Huang, P. Wang,  X. Cheng, D. Zhou, Q. Geng, and R. Yang are with Baidu Research.}% <-this % stops an unwanted space
\thanks{* Equal contribution}}

\IEEEtitleabstractindextext{%
\begin{abstract}
Autonomous driving has attracted tremendous attention especially in the past few years.
The key techniques for a self-driving car include solving tasks like 3D map construction, self-localization, parsing the driving road and understanding objects, which enable vehicles to reason and act. However, large scale data set for training and system evaluation is still a bottleneck for developing robust perception models.
% Various tasks are necessary towards solving self-driving such as semantic 2D parsing, road lane mark parsing, object instance detection and segmentation, 3D self-localization.
In this paper, we present the \emph{ApolloScape} dataset~\cite{apolloweb} and its applications for autonomous driving.
Compared with existing public datasets from real scenes, \eg~KITTI~\cite{Geiger2013IJRR} or Cityscapes~\cite{Cordts2016Cityscapes}, ApolloScape contains much large and richer labelling including holistic semantic dense point cloud for each site, stereo, per-pixel semantic labelling, lanemark labelling, instance segmentation, 3D car instance, high accurate location for every frame in various driving videos from multiple sites, cities and daytimes.
For each task, it contains at lease 15x larger amount of images than SOTA datasets. To label such a complete dataset, we develop various tools and algorithms specified for each task to accelerate the labelling process, such as joint 3D-2D segment labeling, active labelling in videos etc.
Depend on \emph{ApolloScape}, we are able to develop algorithms jointly consider the learning and inference of multiple tasks. In this paper, we provide a sensor fusion scheme integrating camera videos, consumer-grade motion sensors (GPS/IMU), and a 3D semantic map in order to achieve robust self-localization and semantic segmentation for autonomous driving.
%Specifically, we first have an initial coarse camera pose obtained from consumer-grade GPS/IMU, based on which a label map can be rendered from the 3D semantic map. Then, the rendered label map and the RGB image are jointly fed into a pose CNN, yielding a corrected camera pose. In addition, to incorporate temporal information, a multi-layer recurrent neural network (RNN) is further deployed improve the pose accuracy.
%Finally, based on the pose from RNN, we render a new label map, which is fed together with the RGB image into a segment CNN which produces per-pixel semantic label.
We show that practically, sensor fusion and joint learning of multiple tasks are beneficial to achieve a more robust and accurate system.
We expect our dataset and proposed relevant algorithms can support and motivate researchers for further development of multi-sensor fusion and multi-task learning in the field of computer vision.

\end{abstract}
}

\maketitle

\IEEEdisplaynontitleabstractindextext

% \fi
\IEEEpeerreviewmaketitle

%%%%%%%%% BODY TEXT
\IEEEraisesectionheading{\section{Introduction}\label{sec:introduction}}
\IEEEPARstart{A} successful self-driving vehicle that is widely applied must include three essential components. Firstly, understanding the environment, where commonly a 3D semantic HD map at the back-end precisely recorded the environment.
Secondly, understanding self-location, where an on-the-fly self-localization system puts the vehicles accurately inside the 3D world, so that it can plot a path to every target location. Thirdly, understanding semantics in the view, where a 3D perceptual system detects other moving objects, guidance signs and obstacles on the road, in order to avoid collisions and perform correct actions.  The prevailing approaches for solving those tasks from self-driving companies are mostly dependent on LIDAR~\cite{velodyne}, whereas vision-based approaches, which have potentially very low-cost,  are still very challenging and under research.
It requires solving tasks such as learning to do visual 3D scene reconstruction~\cite{kar2017learning,huang2018deepmvs,Yao_2018_ECCV,cheng2018depth}, self-localization~\cite{Kendall_2015_ICCV,schonberger2018semantic}, semantic parsing~\cite{long2015fully,ChenPSA17}, semantic instance understanding~\cite{he2018mask,chen2017masklab}, object 3D instance understanding~\cite{xiang2014beyond,kar2015category,guney2015displets,kundu20183d,songApolloCar3D} online in a self-driving video etc.
However, the SOTA datasets for supporting these tasks either have limited amount, \eg KITTI~\cite{Geiger2013IJRR} only has 200 training images for semantic understanding, or limited variation of tasks, \eg Cityscapes~\cite{Cordts2016Cityscapes} only has discrete semantic labelled frames without tasks like localization or 3D reconstruction.
Therefore, in order to have a holistic training and evaluation of a vision-based self-driving system, in this paper, we build the \emph{Apolloscape}~\cite{apolloweb} for autonomous driving, which is a growing and unified dataset extending previous ones both on the data scale, label density and variation of tasks.

\begin{figure*}[t]
  \centering
  \includegraphics[width=\linewidth]{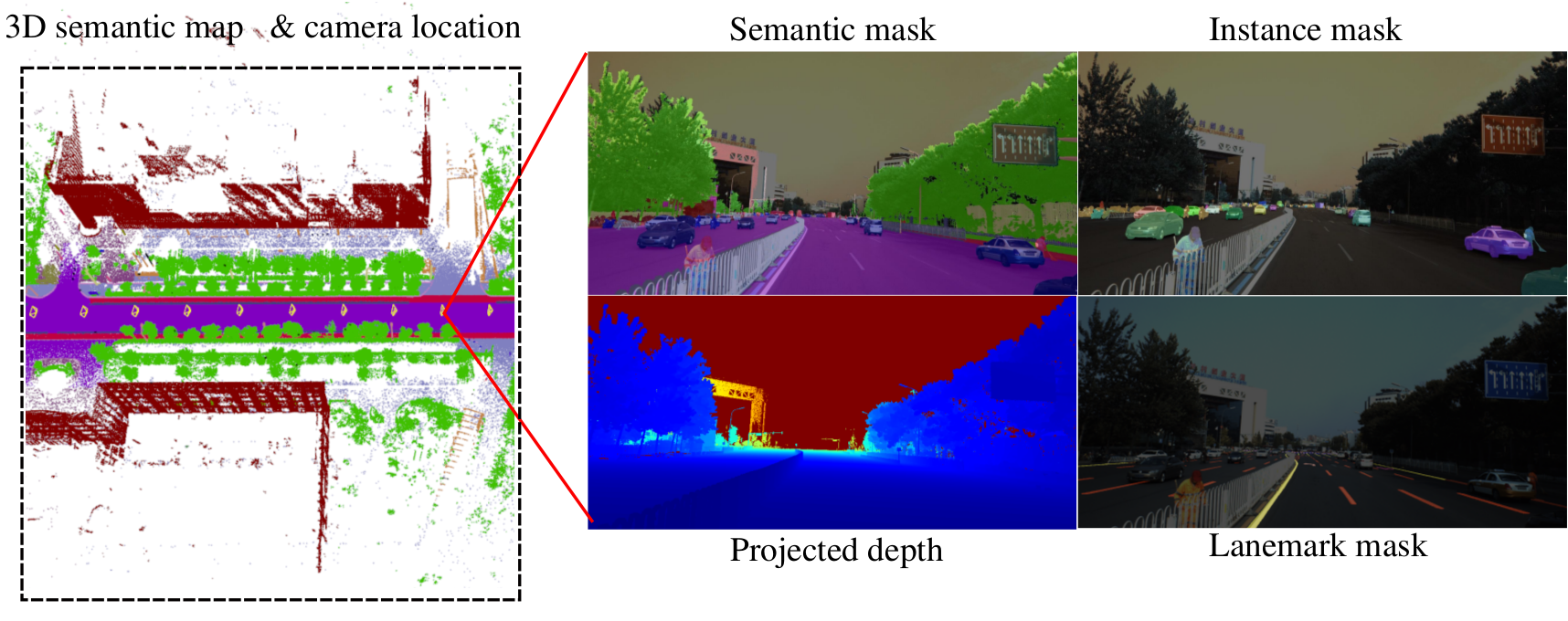}
  \caption{A glance of ApolloScape with various properties. The images are cropped for better visualization. } % (a) 3D semantic point cloud, (b) color image, (c) 2D semantic label, (d) 2D instance label, (e) camera pose and rendered background depth map accordingly. (f) lanemark segmentation. (g) labelled 2D car keypoints and 3D car instances.}
  \label{fig:example}
  \vspace{-1\baselineskip}
\end{figure*}
%At bottom, we show the road map summarizing the past and future of ApolloScape.

Specifically, in current stage, ApolloScape contains properties of,
\begin{enumerate}
    \item dense semantics 3D point cloud for the environment (20+ driving site)
    \item stereo driving videos (100+ hours)
    \item high accurate 6DoF camera pose. (translation $\leq 50$mm, rotation $\leq 0.015^\circ$)
    \item videos at same site under different day times, (morning, noon, night)
    \item dense per-pixel per-frame semantic labelling (35 classes, 144K+ images)
    \item per-pixel lanemark labelling (35 classes, 160K+ images)
    \item semantic 2D instances segmentation (8 classes, 90K+ images)
    \item 2D car keypoints and 3D car instance labelling (70K cars)
\end{enumerate}

With these information, we have released several standard benchmarks for scene parsing~\cite{segment}, instance segmentation~\cite{instance}, lanemark parsing~\cite{lanemark}, self-localization~\cite{localization}  by withholding part of the data as test set, and our toolkit for visualization and evaluation has also published~\cite{apolloapi}. Here, for 3D car instance, we list the car number we already labelled, and since it is still under development, we will elaborate it in our future work.
Fig.~\ref{fig:example} shows a glance of \emph{ApolloScape}, which illustrates various information from the dataset that is necessary for autonomous driving. Our dataset is still growing and evolving, and will shortly contains new tasks such as 3D car instance shape and pose, 3D car tracking etc., which are important for scene understanding with finer granularity. In addition, thanks to our efficient labelling pipeline, we are able to scale the dataset to multiple cities and sites, and we have already contained 10 cities in China  under various driving conditions. % such as raining, snowing and foggy.
%More excitingly, we show a road map of ApolloScape at the bottom of Fig.~\ref{fig:example}.

Based on \emph{ApolloScape}, we are able to develop algorithms for jointly considering 3D and 2D simultaneously with multiple tasks like segmentation, reconstruction, self-localization etc. These tasks are traditionally handled individually~\cite{Kendall_2015_ICCV,ChenPSA17}, or jointly handled offline with semantic SLAM~\cite{kundu2014joint} which could be time consuming.
However, from a more practical standpoint, self-driving car needs to handle localization and parsing the environment on-the-fly efficiently. Therefore, in this paper, we propose a deep learning based online algorithm jointly solving localization and semantic scene parsing when a 3D semantic map is available. In our system, we assume to have (a) GPS/IMU signal to provide a coarse camera pose estimation; (b) a semantic 3D map for the static environment. The GPS/IMU signals serve as a crucial prior for our pose estimation system. The semantic 3D map, which can synthesize a semantic view for a given camera pose, not only provides strong guidance for scene parsing, but also helps maintain temporal consistency.

With our framework, the camera poses and scene semantics are mutually beneficial. The camera poses help establish the correspondences
between the 3D semantic map and 2D semantic label map. Conversely, scene semantics could help refine camera poses. Our unified framework yields better results, in terms of both accuracy and speed, for both tasks than doing them individually. In our experiments, using a single Titan Z GPU, the networks in our system estimates the pose in 10ms with accuracy under 1 degree, and segments the image $\by{512}{608}$ within 90ms with pixel accuracy around 96$\%$ without model compression, which demonstrates its
efficiency and effectiveness.

In summary, the contributions of this work are in three folds,
\begin{enumerate}
  % \item The first subset, 143,906 image frames with pixel annotations, has been released. We divide our dataset into easy, moderate, and hard subsets. The difficulty levels are measured based on number of vehicles and pedestrians per image that often indicates the scene complexity. Our goal is to capture and annotate around one million video frames and corresponding 3D point clouds.
  % \item Our dataset has survey-grade dense 3D point cloud for static objects. A rendered depth map is associated with each image, creating the first pixel-annotated RGB-D video for outdoor scenes.
  % \item In addition to typical object annotations, our dataset also contains fine grain labelling of lane markings (with 28 classes).
  \item We propose a large and rich dataset, named as \emph{ApolloScape}, which includes various tasks, \eg 3D reconstruction, self-localization, semantic parsing, instance segmentation etc., supporting the training and evaluation of vision-based autonomous driving algorithms and systems.
  \item For developing the dataset, we design an efficient and scalable 2D/3D joint-labelling pipeline, where various tools are developed for 2D segmentation, 3D instance understanding etc. For example, compared with fully manual labelling, our 3D/2D labelling pipeline saves 70\% labeling time for semantic segmentation.
  % Based on our labelling pipeline, all the 3D point clouds will be assigned with above annotations. Therefore, our dataset is the first open dataset of street views containing 3D annotations.
  \item Based on \emph{ApolloScape}, we developed a deep learning based joint self-localization and segmentation algorithm, which is relying on a semantic 3D map. The system fuses sensors from camera and customer-grad GPS/IMU, which runs efficiently and improves the robustness and accuracy for camera localization and scene parsing. % The instance-level annotations are available for video frames, which are especially useful to design spatial-temporal models for prediction, tracking, and behavior analysis of movable objects.
\end{enumerate}

The structure of this paper is organized as follows. We provide related work in Sec.~\ref{sec:related}, and elaborate the collection and labelling of \emph{ApolloScape} in Sec.~\ref{sec:dataset}. In Sec.~\ref{sec:localize_and_parsing}, we explain the  developed efficient joint segmentation and localization algorithm. Finally, we present the evaluation results of our algorithms, the benchmarks for multiple tasks and corresponding baseline algorithms performed on these tasks in Sec.~\ref{sec:experiments}.

\section{Related works}\label{sec:related}

Autonomous driving datasets and related algorithms has been an active research area for years.
% Estimating camera pose and semantic parsing given a
% video or a single image have long been center problems for
% computer vision.
Here we summarize the related works in aspects of datasets and most relevant algorithms without enumerating them all due to space limitation.

\begin{table*}[!htpb]
  \centering
  \begin{threeparttable}
  \begin{tabular}{ l c c c c c c c}
     \toprule
     %Dataset & Geo Accuracy & Diversity & \multicolumn{4}{c}{Annotation} \\
     %\multirow{1}{*}
     %\multicolumn{3}{l}{} & 3D & 2D & Video & Lane\\

     \multicolumn{1}{l}{\multirow{2}{*}{Dataset}} & \multicolumn{1}{l}{\multirow{2}{*}{Real}} & \multicolumn{1}{c}{\multirow{2}{*}{Location Accuracy}} &
     \multicolumn{1}{c}{\multirow{2}{*}{Diversity}} & \multicolumn{4}{c}{Annotation} \\
     \cmidrule{5-8}
     \multicolumn{4}{c}{} & 3D & 2D & Video & Lane\\

     \midrule
     CamVid~\cite{brostow2009semantic} & \checkmark & - & day time & no & pixel: 701 &  \checkmark & 2D / 2 classes\\
     \midrule
     Kitti~\cite{Geiger2013IJRR} & \checkmark & cm & day time & 80k 3D box & box: 15k & - & no \\
      &  & & &  & pixel: 400 &  &  \\
     \midrule
     Cityscapes~\cite{Cordts2016Cityscapes} & \checkmark & - & day time & no & pixel: 25k & - & no \\
      & &  & 50 cities & & & &\\
     \midrule
     Toronto~\cite{wang2017torontocity} & \checkmark & cm & Toronto &  \multicolumn{4}{c}{focus on buildings and roads}\\
     & &  &  &  \multicolumn{4}{c}{exact numbers are not available\tnote{1}}\\
     \midrule
     Mapillary~\cite{neuhold2017mapillary} & \checkmark & meter & various weather & no & pixel: 25k & - & 2D / 2 classes \\
      & & & day \& night & & & &\\
      & & & 6 continents & & & &\\
     \midrule
     BDD100K~\cite{yu2018bdd100k} & \checkmark & meter & various weather & no & box: 100k & - & 2D / 2 classes \\
                                  &       & & day             &    & pixel: 10k& &\\
                                  &       & & 4 regions in US & &  & &\\
     \midrule
     SYNTHIA~\cite{RosCVPR16}     & - & - & various weather & box &  pixel:213k & \checkmark & no \\
     \midrule
     P.F.B.~\cite{richter2017playing} & - & - & various weather & box &  pixel:250k  & \checkmark & no \\
     \midrule
     \midrule
     ApolloScape & \checkmark & cm & various weather & 3D semantic point & pixel: 140k & \checkmark & 3D / 2D Video \\
                 & & & day time & 70K 3D fitted cars &  & & 27 classes \\
                 & & & 4 regions in China &  &   &  & \\
     \bottomrule
   \end{tabular}
   \begin{tablenotes}
    \item[1] database is not open to public yet.
   \end{tablenotes}
   \end{threeparttable}
     \caption{Comparison between our dataset and the other street-view self-driving datasets published. ``pixel" represents 2D pixel-level annotations. ``point" represents 3D point-level annotations. ``box" represents bounding box-level annotations. ``Video" indicates whether 2D video sequences are annotated. ``3D fitted cars'' gives the number of car instance we already fitted in the images with a 3D mesh model, which we will introduce in our future works.}
     \label{tb:sum}
     \vspace{-1.5\baselineskip}
\end{table*}

\subsection{Datasets for autonomous driving.}
Most recently, various datasets targeting at solving each individual visual task for robot navigation have been released such as 3D geometry estimation~\cite{scharstein2014high,silberman2012indoor},  localization~\cite{Kendall_2015_ICCV,sattler2018benchmarking}, instance detection and segmentation~\cite{everingham2010pascal,lin2014microsoft}.
However, focusing on autonomous driving, a set of comprehensive visual tasks are preferred to be collected consistently within a unified dataset from driving videos, so that one may explore the mutual benefits between different problems.

In past years, lots of datasets have been collected in various cities, aiming to increase variability and complexity of urban street views for self-driving applications.
The Cambridge-driving Labeled Video database (CamVid)~\cite{brostow2009semantic} is the first dataset with semantic annotated videos. The size of the dataset is small, containing 701 manually annotated images with 32 semantic classes.
The KITTI vision benchmark suite~\cite{Geiger2013IJRR} is later collected and contains multiple computer vision tasks such as stereo, optical flow, 2D/3D object detection and tracking.
For semantics, it mainly focuses on detection, where 7,481 training and 7,518 test images are annotated by 2D and 3D bounding boxes, and each image contains up to 15 cars and 30 pedestrians. Nevertheless, for segmentation, very few images contain pixel-level annotations, yielding a relatively weak benchmark for semantic segmentation.
Most recently, the Cityscapes dataset~\cite{Cordts2016Cityscapes} is specially collected for 2D segmentation which contains 30 semantic classes. In detail, 5,000 images have detailed annotations, and 20,000 images have coarse annotations. Although video frames are available, only one image out of each video is manually labelled. Thus, tasks such as video segmentation can not be performed.
Similarly, the Mapillary Vistas dataset~\cite{neuhold2017mapillary} provides a larger set of images with fine annotations, which has 25,000 images with 66 object categories.
The TorontoCity benchmark~\cite{wang2017torontocity} collects LIDAR data and images including stereo and panoramas from both drones and moving vehicles. Although the dataset scale is large, which covers the Toronto area. as mentioned by authors, it is not possible to manually do per-pixel labelling of each frame. Therefore, only two semantic classes, i.e., building footprints and roads, are provided for benchmarks of segmentation.
BDD100K database~\cite{yu2018bdd100k} contains 100K raw video sequences representing more than 1000 hours of driving hours with more than 100 million images. Similarly with the Cityscapes, one image is selected from each video clip for annotation. 100K images are annotated in bounding box level and 10K images are annotated in pixel level.

Real data collection is laborious, to avoid the difficulties in real scene collection, several synthetic datasets are also proposes. SYNTHIA~\cite{RosCVPR16} builds a virtual city with Unity development platform~\cite{unity}, and Play for benchmark~\cite{richter2017playing} extracts ground truth with GTA game engine. Though large amount of data and ground truth can be generated, there is still a domain gap~\cite{hoffman2016fcns} between appearance of synthesized images and the real ones. In general, models learned in real scenario still generalize better in real applications such as object detection and segmentation~\cite{zhang2017curriculum,chen2018road}.

In Tab.~\ref{tb:sum}, we compare the properties our dataset and other SOTA datasets for autonomous driving, and show that \emph{ApolloScape} is unique in terms of data scale, granularity of labelling, task variations within real environments. Later in Sec.~\ref{sec:dataset}, we will present more details about the dataset.

\subsection{Self-localization and semantic scene parsing.}
As discussed in Sec.~\ref{sec:introduction}, we also try to tackle real-time self-localization and semantic scene parsing back on \emph{ApolloScape} given a video or a single image. These two problems have long been center focus for computer vision. Here we summarize the related works on outdoor cases with street-view images as input.

% talk each perspective with image and video
\medskip
\noindent\textbf{Visual self-localization.} Traditionally, localizing an image given a set of 3D points is formulated as a Perspective-$n$-Point (P$n$P) problem~\cite{haralick1994review,kneip2014upnp} by matching feature points in 2D and features in 3D through cardinality maximization. Usually in a large environment, a pose prior is required in order to obtain good estimation~\cite{david2004softposit,moreno2008pose}. Campbell \etal~\cite{campbell2017globally} propose a global-optimal solver which leverage the prior. In the case that geo-tagged images are available, Sattler \etal~\cite{sattler2017large} propose to use image-retrieval to avoid matching large-scale point cloud.
When given a video, temporal information could be further modeled with methods like SLAM~\cite{engel2014lsd} etc, which increases the localization accuracy and speed.

Although these methods are effective in cases with distinguished feature points, they are still not practical for city-scale environment with billions of points, and they may also fail in areas with low texture, repeated structures, and occlusions.
Thus, recently, deep learned features with hierarchical representations are proposed for localization. PoseNet~\cite{Kendall_2015_ICCV,kendall2017geometric} takes a low-resolution image as input, which can estimate pose in 10ms \wrt a feature rich environment composed of distinguished landmarks.
LSTM-PoseNet~\cite{hazirbasimage} further captures a global spatial context after CNN features.
Given an video, later works incorporate Bi-Directional LSTM~\cite{DBLP:journals/corr/ClarkWMTW17} or Kalman filter LSTM~\cite{coskun2017long} to obtain better results with temporal information. Most recently, many works~\cite{schonberger2018semantic,lianos2018vso} also consider adding semantic cues as more robust representation for localization.  However, in street-view scenario, considering a road with trees aside, in most cases, no significant landmark appears, which could fail the visual models. Thus, signals from GPS/IMU are a must-have for robust localization in these cases~\cite{vishal2015accurate}, whereas the problem switched to estimating the relative pose between the camera view from a noisy pose and the real pose. For finding relative camera pose of two views, recently, researchers~\cite{laskar2017camera,ummenhofer2016demon} propose to stack the two images as a network input.
In our case, we concatenate the real image with an online rendered label map from the noisy pose, which provides superior results in our experiments.
\medskip
\noindent\textbf{Street scene parsing.} For parsing a single image of street views (e.g., these from CityScapes~\cite{Cordts2016Cityscapes}), most state-of-the-arts (SOTA) algorithms are designed based on a FCN~\cite{long2015fully} and a multi-scale context module with dilated convolution~\cite{ChenPSA17}, pooling~\cite{ZhaoSQWJ16}, CRF~\cite{higherordercrf_ECCV2016}, or spatial RNN~\cite{byeon2015scene}. However, they are dependent on a ResNet~\cite{he2016deep} with hundreds of layers, which is too computationally expensive for applications that require real-time performance.
Some researchers apply small models~\cite{PaszkeCKC16} or model compression~\cite{ZhaoQSSJ17} for acceleration, with the cost of reduced accuracy.
When the input is a video, spatial-temporal informations are jointly considered, Kundu \etal~\cite{kundu2016feature} use 3D dense CRF to get temporally consistent results. Recently, optical flow~\cite{dosovitskiy2015flownet} between consecutive frames is computed to transfer label or features~\cite{gadde2017semantic,zhu2016deep} from the previous frame to current one.  In our case, we connect consecutive video frames through 3D information and camera poses, which is a more compact representation for static background.
In our case, we propose the projection from 3D maps as an additional input, which alleviates the difficulty of scene parsing solely from image cues. Additionally, we adopt a light weighted network from DeMoN~\cite{ummenhofer2016demon} for inference efficiency.%\textcolor[rgb]{1.00,0.00,0.00}{of what?}.

\medskip
\noindent\textbf{Joint 2D-3D for video parsing.} Our work is also related to joint reconstruction, pose estimation and parsing~\cite{kundu2014joint,hane2013joint} through embedding 2D-3D consistency.
 Traditionally, reliant on structure-from-motion (SFM)~\cite{hane2013joint} from feature or photometric matching, those methods first reconstruct a 3D map, and then perform semantic parsing over 2D and 3D jointly, yielding geometrically consistent segmentation between multiple frames.
 Most recently, CNN-SLAM~\cite{tateno2017cnn} replaces traditional 3D reconstruction module with a single image depth network, and adopts a segment network for image parsing.
 However, all these approaches are processed off-line and only for static background, which do not satisfy our online setting. Moreover, the quality of a reconstructed 3D model is not comparable with the one collected with a 3D scanner. % \textcolor[rgb]{1.00,0.00,0.00}{you mean our scanner?} 

\section{Build \emph{ApolloScape}}\label{sec:dataset}
In this section, we introduce our acquisition system, specifications about the collected data and efficient labelling process for building ApolloScape.

\subsection{Acquisition system}\label{subsec:acq}
\begin{figure}[t]
  \centering
  \includegraphics[width=\linewidth]{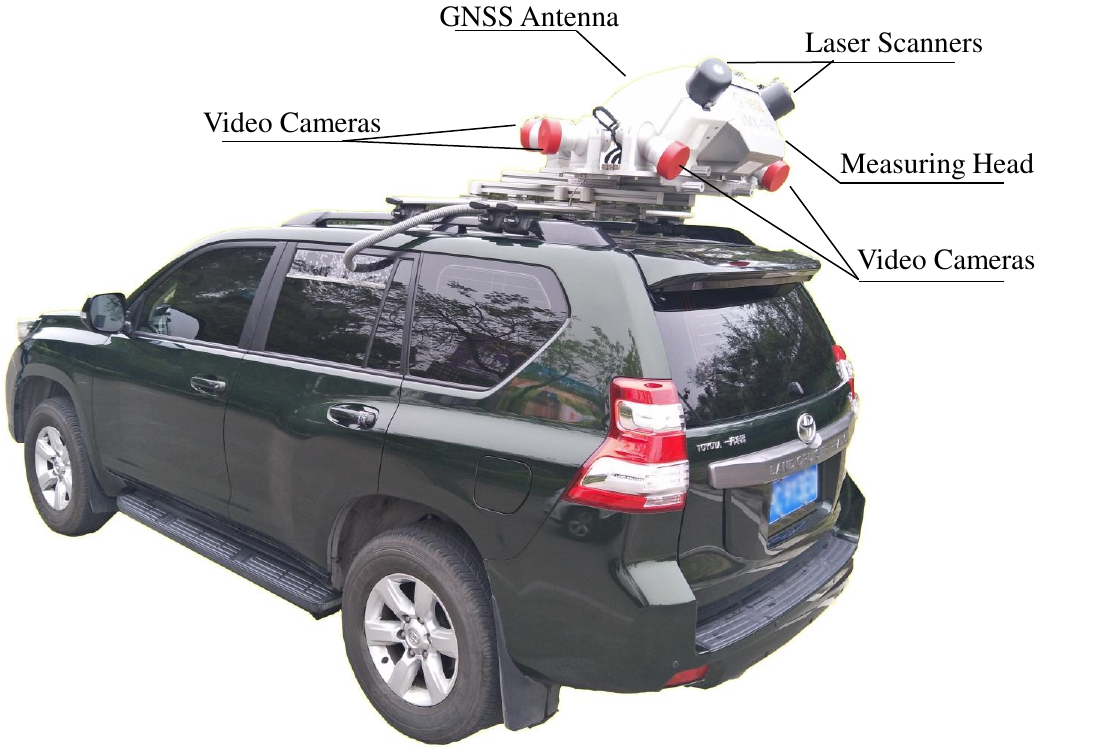}
  \caption{Acquisition system consists of two laser scanners, up to six video cameras, and a combined IMU/GNSS system.}\label{fig:platform}
\end{figure}

In Fig.~\ref{fig:platform}, we visualize our collection system.
To collect static 3D environment, we adopt Riegl VMX-1HA~\cite{rieglvmx} as our acquisition system that consists of two VUX-1HA laser scanners ($360^\circ$ FOV, range from 1.2m up to 420m with target reflectivity larger than 80\%), one VMX-CS6 camera system (two front cameras are used with resolution $3384\times2710$), and a measuring head with IMU/GNSS (position accuracy $20\sim50$mm, roll \& pitch accuracy $0.005^\circ$, and heading accuracy $0.015^\circ$).
The laser scanners utilizes two laser beams to scan its surroundings vertically that are similar to the push-broom cameras. Comparing with common-used Velodyne HDL-64E~\cite{velodyne}, the scanners are able to acquire higher density of point clouds and obtain higher measuring accuracy / precision (5mm / 3mm). The whole system has been internally calibrated and synchronized, and is mounted on the top of a mid-size SUV.

Additionally, the system contains two high frontal camera capturing with a resolution of \by{3384}{2710}, and is well calibrated with the LIDAR device. Finally, to obtain high accurate GPS/IMU information, a temporary GPS basement is set up near the collection site to make sure the localization of the camera is sufficiently accurate for us to match the 2D image and 3D point cloud.
Commonly, our vehicle drives at the speed of 30km per hour and the cameras are triggered once every meter, \ie 30fps.

% However, the acquired point clouds of moving objects could be highly distorted or even completely missing.
% As an ongoing project, we plan to mount stereo cameras and a panoramic camera system in near future to further produce complete depth information and panoramic images.

\begin{table}
  \centering
  \scalebox{1}{
  \begin{tabular}{ l c c c c c c}
     \toprule
     Count & Kitti & Cityscapes & BDD100K & \multicolumn{3}{c}{ApolloScape} \\
     & (box) & (pixel) & (box) & \multicolumn{3}{c}{(pixel)} \\
     \midrule
     \multicolumn{6}{l}{\textbf{total $(\times10^4)$}}\\
     \midrule
     person & 0.6 & 2.4 & 12.9 & \multicolumn{3}{c}{\textbf{54.3}}  \\
     vehicle & 3.0 & 4.1 & 110.2 & \multicolumn{3}{c}{\textbf{198.9}} \\
     \midrule
     \multicolumn{4}{l}{\textbf{average per image}} & \textbf{e} & \textbf{m} & \textbf{h}\\
     \midrule
     person & 0.8 & 7.0 & 1.3 &  1.1 & 6.2 & \textbf{16.9} \\
     vehicle & 4.1 & 11.8 & 11.0 & 12.7 & 24.0 & \textbf{38.1} \\
    %  \midrule
    %  car & - & - & 10.2 & 9.7 & 16.6 & 24.5 \\
    %  motorcycle & - & - & 0.0 & 0.1 & 0.8 & 2.5 \\
    %  bicycle & - & - & 0.1 & 0.2 & 1.1 & 2.4 \\
    %  rider & - & - & 0.1 & 0.8 & 3.3 & 6.3  \\
    %  truck & - & - & 0.4 & 0.8 & 0.8 & 1.4 \\
    %  bus & - & - & 0.2 & 0.7 & 1.3 & 0.9 \\
    %  tricycle & 0.0 & 0.0 & 0.0 & 0.4 & 0.3 & 0.2 \\
     % train & - & - & 0.0 & 0.0 & 0.0 & 0.0 \\
     \bottomrule
   \end{tabular}
   }
   \caption{\peng{Total and average number of instances} in KITTI~\cite{Geiger2013IJRR}, Cityscapes~\cite{Cordts2016Cityscapes}, BDD100K~\cite{yu2018bdd100k}, and \emph{ApolloScape} (instance-level). ``pixel" represents 2D pixel-level annotations. ``box" represents bounding box-level annotations. The letters, e, m, and h, indicate easy, moderate, and hard subsets in \emph{ApolloScape} respectively.}
   \label{tb:stats1}
   \vspace{-1.5\baselineskip}
\end{table}

\begin{figure*}
  \centering
  \includegraphics[width=\linewidth]{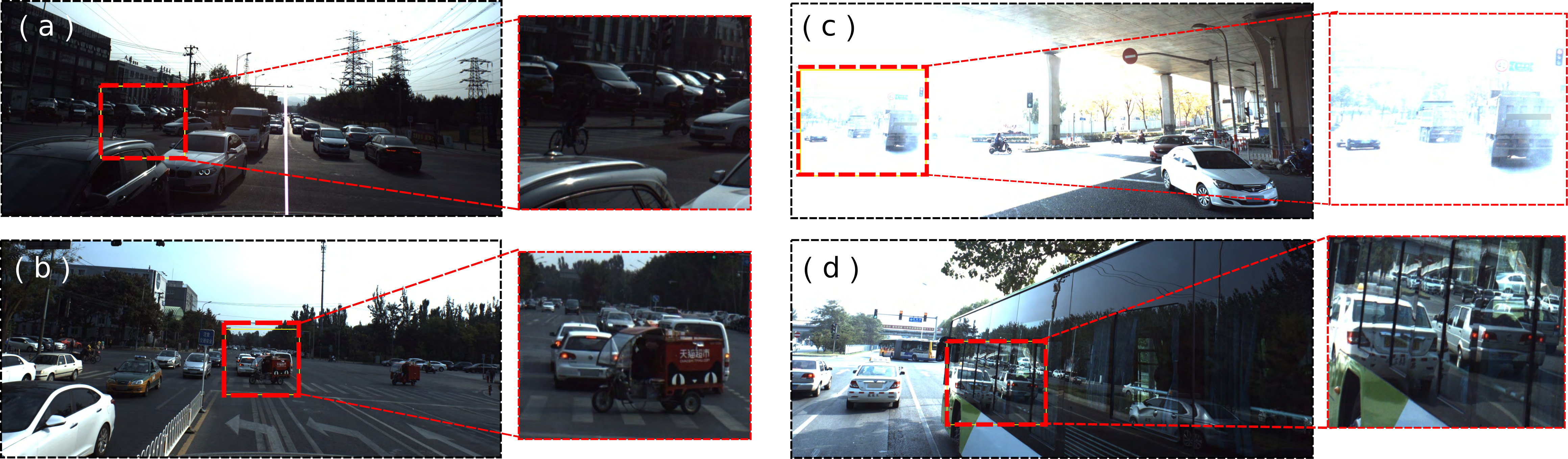}
  \caption{Examples with challenging environments for object detection and segmentation (Images are center-cropped for better visualization). We highlight and zoom in the region of challenges in each image. (a) Objects with heavy occlusion and small scale. (b) Abnormal action by cyclist drivers. (c) High contrast and overexposure due to shadows and strong sunlight. (d) Mirror reflection on bus glasses.
  %The last row contains enlarged regions enclosed by yellow rectangles.
  }
  \label{fig:scene}
   \vspace{-0.8\baselineskip}
\end{figure*}

\subsection{Specifications} \label{subsec:spec}
Here, based on the acquisition system, we first present the specifications of \emph{Apolloscape} \wrt different tasks, \eg predefined semantic classes, lanemark classes and instance etc., to allow better overview of the dataset. In Sec.~\ref{subsec:label}, we will introduce our active labelling pipeline which allows us to efficiently produce the ground truth of multiple tasks simultaneously.
% Notice that, due to occlusions among multiple objects, pixel-level annotations provide more accurate information than bounding box-level annotations in general.

\medskip
\noindent\textbf{Semantic scene parsing.} In our current version released online~\cite{segment,instance}, we have 143,906 video frames and their corresponding pixel-level semantic labelling, from which 89,430 images contain instance-level annotations where movable objects are further separated. Notice that our labelled images contains temporal information which could also be useful for video semantic and object segmentation. % Table~\ref{tb:sum} shows a comparison of several key properties between our dataset and other street-view datasets in real world.

To make the evaluation more comprehensive, similar to the KITTI~\cite{Geiger2013IJRR},
we separate the recorded video with the level of easy, moderate, and heavy scene complexities based on the amount of movable objects, such as person and vehicles.
Tab.~\ref{tb:stats1} compares the scene complexities between \emph{ApolloScape},  the Cityscapes~\cite{Cordts2016Cityscapes} and KITTI~\cite{Geiger2013IJRR}, where we show the statistics for each individual classes of movable objects.
\emph{ApolloScape} contains more objects than others in terms of both total number and average number of object instances from images.
More importantly, it contains stronger challenging environments, as shown in Fig.~\ref{fig:scene}. For instance, high contrast regions due to sun light and large area of shadows from the overpass. Mirror reflections of multiple nearby vehicles on a bus glass due to highly crowded transportation. We hope these case can help and motivate researchers to develop more robust models against environment changes.
% We will continue releasing more data in near future with large diversities of location, traffic conditions, and weathers.

\begin{figure*}[t]
  \centering
  \includegraphics[width=0.95\linewidth]{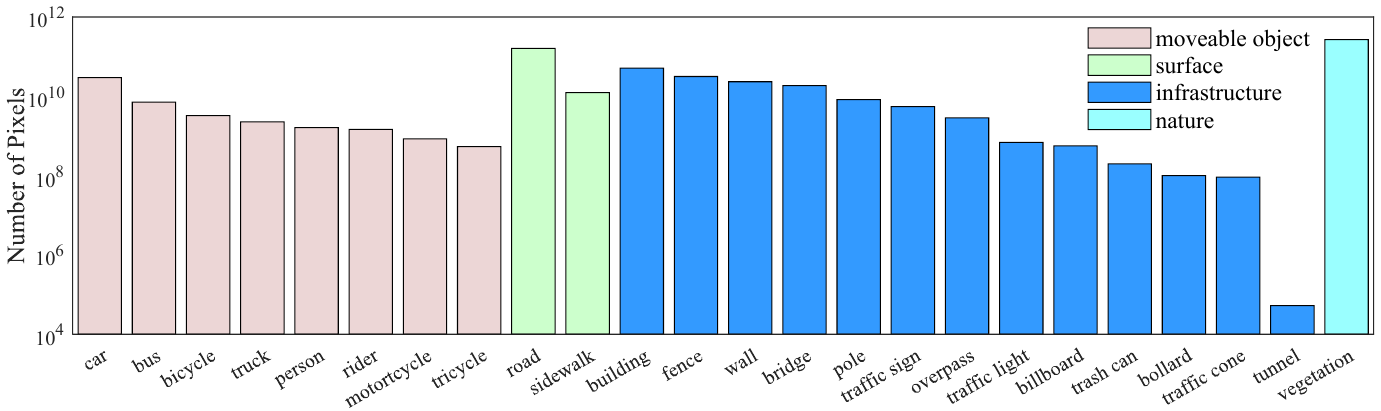}
  \caption{24 semantic classes and corresponding numbers of annotated pixels. Bar colors indicate different semantic groups.
  %The last row contains enlarged regions enclosed by yellow rectangles.
  }
  \label{fig:apollo_counts}
  %\vspace{\baselineskip}
\end{figure*}

For semantic scene parsing, we annotate 24 different labels in four groups. The specifications of the classes are partially borrowed from the Cityscapes dataset. Fig.~\ref{fig:apollo_counts} gives the amount of labelled pixels for each class. As expected, \emph{ApolloScape} provides much higher amount of average annotated pixels than Cityscapes, especially for some rare classes, \eg~\textit{traffic-light, pole}.  Here, we add several new classes common in China. For instance, we add ``tricycle" that is one of the most popular means of transportation. This class covers all kinds of three-wheeled vehicles that could be both motorized and human-powered. The rider class in the Cityscape is defined as the person on means of transportation. Here, we consider the person and the means of transportation as a single moving object, and treat the two together as one class. The three classes related to rider, i.e., bicycle, motorcycle, and tricycle, represent means of transportation without rider and parked along the roads.

%\begin{table}\clearpage
%\centering
%\scalebox{1.2}{
%\begin{tabular}{ p{4.7em} p{4.5em} c p{6.5em} }
%  \toprule
%  % after \\: \hline or \cline{col1-col2} \cline{col3-col4} ...
%  Group & Class & ID & Description \\
%  \midrule
%  movable & car & 1 &  \\
%  object & motorcycle & 2 & \\
%  & bicycle & 3 & \\
%  & person & 4 & \\
%   & rider & 5 & person on \\
%   & & & motorcycle, \\
%   & & & bicycle or \\
%   & & & tricycle\\
%   & truck & 6 & \\
%   & bus & 7 &   \\
%   & tricycle & 8 & three-wheeled \\
%   & & & vehicles, \\
%   & & & motorized, or \\
%   & & & human-powered \\
%  \midrule
%  surface & road & 9 &  \\
%   & sidewalk & 10 &  \\
%  \midrule
%  infrastructure & traffic cone & 11 & movable and \\
%  & & & cone-shaped markers\\
%  & bollard & 12 & short,vertical post \\
%  & fence & 13 &  \\
%  & traffic light & 14 & \\
%  & pole & 15 & \\
%  & traffic sign & 16 & \\
%  & wall & 17 & \\
%  & trash can & 18 & \\
%  & billboard & 19 & \\
%  & building & 20 & \\
%  & bridge & 255 & \\
%  & tunnel & 255 & \\
%  & overpass & 255 & \\
%  \midrule
%  nature & vegetation & 21 & \\
%  \midrule
%  void & void & 255 & unlabeled regions \\
%  \bottomrule
%\end{tabular}
%}
%\caption{Details of semantic classes in ApolloScape, where our unique classes (not appeared in Cityscapes~\cite{Cordts2016Cityscapes}) are described.}
%\label{tb:class1}
%\vspace{-0.3\baselineskip}
%\end{table}

\begin{figure*}[!htpb]
  \centering
  \includegraphics[width=0.95\linewidth]{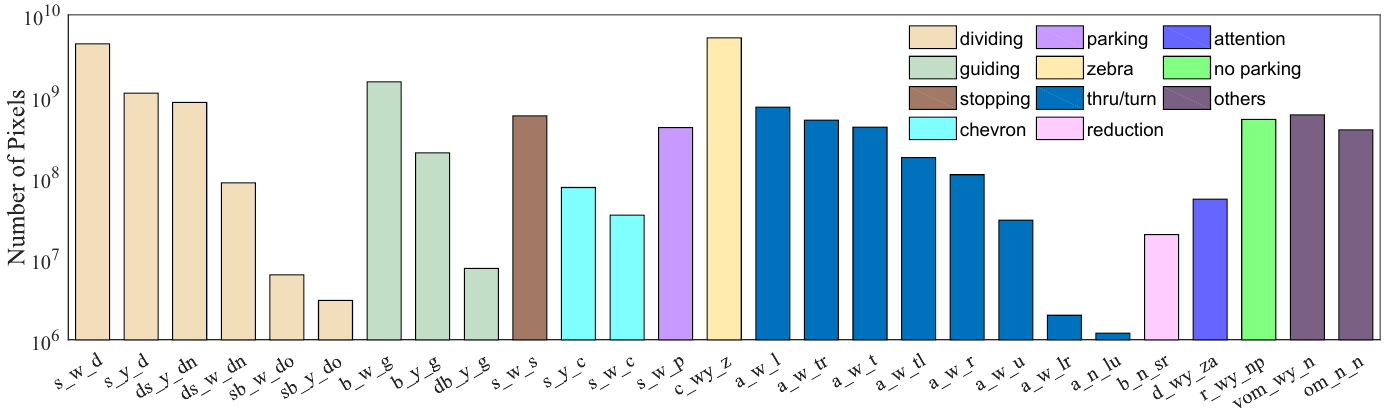}
  \caption{27 lane mark labels and corresponding numbers of annotated pixels. Bar colors indicate 11 different lane mark usages. Here,``s\_w\_d" is short for solid, white and dividing in Tab.~\ref{tb:class2} by combining the first letter of type, color and usage respectively, and other classes are named accordingly.
  %The last row contains enlarged regions enclosed by yellow rectangles.
  }
  \label{fig:lane_counts}
  \vspace{-1\baselineskip}
\end{figure*}

\medskip
\noindent\textbf{Semantic lanemark segmentation.} Automatically understanding lane mark is perhaps the most important function for autonomous driving since it is the guidance for possible actions.
In \emph{ApolloScape}, 27 different lane markings are used for evaluation as elaborated in Tab.~\ref{tb:class2} and Fig.~\ref{fig:lane_counts}. The labels are defined based on lane mark attributes including color (e.g., white and yellow) and type (e.g., solid and broken). To be specific, 165949 images from 3 road sites are labelled and released online~\cite{lanemark}, where 33760 images are withheld for testing.  Comparing to other public available datasets such as KITTI~\cite{Geiger2013IJRR} or the one from Tusimple~\cite{lanemark_tu}, \emph{ApolloScape} is the first large dataset containing rich semantic labelling for lane marks with many variations. %In our future release, we will have additional rich labelling ann.
% Table~\ref{tb:class2} gives detailed information of these lane markings.

\begin{table}
\centering
\scalebox{1.1}{
\begin{tabular}{ l c l }
  \toprule
  Type & Color & Use\\
  \midrule
  solid & w & dividing\\
  solid & y & dividing\\
  double solid  & w & dividing, no pass\\
  double solid  & y & dividing, no pass\\
  solid \& broken & y & dividing, one-way pass\\
  solid \& broken & w & dividing, one-way pass\\
  \midrule
  broken & w & guiding\\
  broken & y & guiding\\
  double broken  & y & guiding\\
  % double broken  & w & guiding\\
  \midrule
  %double broken & w & stopping\\
  %double solid & w & stopping\\
  solid & w & stopping\\
  \midrule
  solid & w & chevron\\
  solid & y & chevron\\
  \midrule
  solid & w & parking\\
  %solid & n/a & parking\\
  \midrule
  % crosswalk & w & parallel & 215 \\
  crosswalk & w & zebra\\
  \midrule
  arrow & w & u-turn\\
  arrow & w & thru\\
  arrow & w & thru \& left turn \\
  arrow & w & thru \& right turn\\
  %arrow & w & thru \& left \& right turn \\
  arrow & w & left turn \\
  arrow & w & right turn\\
  arrow & w & left \& right turn \\
  arrow & w & left \& u-turn \\
  %arrow & w & thru \& u-turn \\
  %arrow & w & merge \\
  %arrow & y & thru \\
 % \midrule
  %$ symbol & w & restricted & 212 \\
  \midrule
   bump & n/a & speed reduction \\
  \midrule
   diamond & w/y & zebra attention \\
  \midrule
   rectangle & w/y & no parking \\
  \midrule
  visible old marking & y/w & others \\
  other markings & n/a & others  \\
  \bottomrule
\end{tabular}
}
\caption{Details of lane mark labels in our dataset (y: yellow, w:white).}\label{tb:class2}
\vspace{-1.5\baselineskip}
\end{table}

\medskip
\noindent\textbf{Self-localization.} Each frame of our recorded video is tagged with high accurate GPS/IMU signal automatically. Therefore, the videos we released for segmentation are also available for self-localization research. However, for setting up a benchmark, we additionally collected a much larger amount of videos, which has not been semantically labelled. %Therefore,
%most recently, we prepare another large set of self-localization
Specifically, videos for localization, as published at~\cite{localization}, contain 6 more roads at 4 different cities, which include roughly 300k images, and road of 28$km$. % In addition, we also uploaded all the dense point cloud related to each road to support approaches requiring 3D map for localization.

Our dataset has variations under different lighting, \ie morning, noon and night, and driving conditions, \ie rush and non-rush hours, with stereo pair of images available. In addition, each road has a  survey-grade point cloud based 3D map that can be used in finding matching pixels for both supervised and unsupervised feature learning~\cite{revaud2016deepmatching,luo2018every} etc.
Finally, we record each road by driving from start-to-end and then end-to-start, which means each position along a road will be looked at from two opposite directions. This enables the research of camera localization with large view changes such as that proposed in semantic visual localization~\cite{schonberger2018semantic}.

% Finally, we will have all our point cloud semantically labelled, and then record additional test sequences on the sites we have already collected for segmentation in order to support learning and fusion of  multitask models.

% \noindent\textbf{3D car understanding.} Finally, to support 3D object understanding with a more detailed granularity, as illustrated in Fig.~\ref{fig:example}, we fit 3D car models for each car instance inside an image, which jointly consider 3D car poses and car shapes. In our current version, 5190 images have 3D car fitted with one of 79 car models. This is 10x larger than similar dataset such as PASCAL 3D~\cite{} and KITTI displets~\cite{}.
% To efficiently label each car, rather than rotating and permuting a car model like the tool provided by PASCAL 3D, we try to label the geometrical key points of a car instance, \eg light corners, door hands. In the meantime, we also have corresponding key points marked on each of our 3D car models. Then we can optimize a PnP~\cite{} problem with the visible key points to obtain a best fitted shape and model from our model set.
% However, it is possible that the amount of key points are not sufficient enough for a robust estimation, Specifically, we consider

% \begin{align}
%     \min_i\min_\ve{T}\|\ve{K}\ve{T}\ve{X}_i - \ve{x}\|_2
% \end{align}
% where $\ve{T}$ is a transformation matrix from  and
% Due to the reason that all our cars are from realistic car models with absolute scales, thus the scale confusion is

\subsection{Labeling Process}\label{subsec:label}
\begin{figure*}[th]
  \centering
  \includegraphics[width=0.92\linewidth]{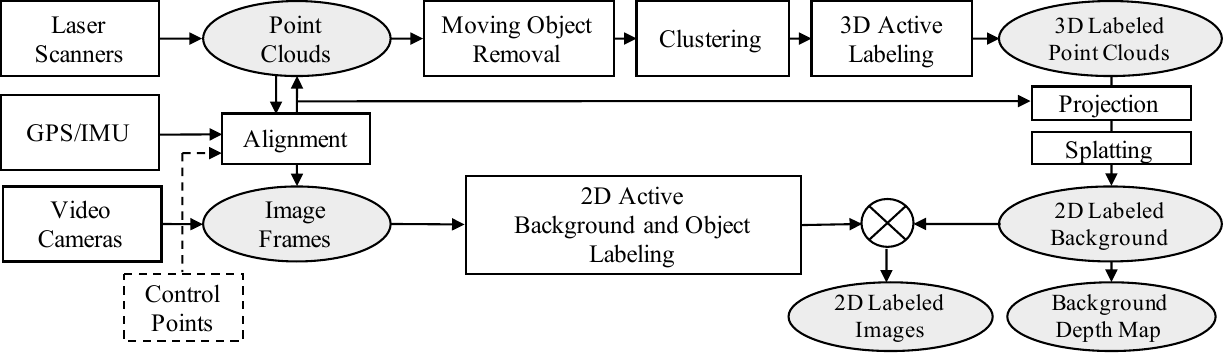}
  \caption{Our 2D/3D labeling pipeline that label static background/objects and moving objects separately. We also adopt active strategies for accelerating the labelling process for scalability of the labelling process. For inputs, since GPS/IMU still has some errors, we manually add control points to better align the point clouds and our image frames. In Sec.~\ref{subsec:label}, we present the details of each components. }
  \label{fig:labeling}
  \vspace{-0.5\baselineskip}
\end{figure*}

In order to make our labeling of video frames accurate and efficient, we propose an active labelling pipeline by jointly consider 2D and 3D information, as shown in Fig.~\ref{fig:labeling}.
The pipeline mainly consists of two stages, 3D labeling and 2D labeling, to handle static background/objects and moving objects respectively.
The basic idea of our pipeline is similar to the one described in~\cite{xie2016semantic}, which transfers the 3D labelled results to 2D images by camera projection, while we need to handle much larger amount of data and have different set up of the acquisition vehicle. Thus some key techniques used in our pipeline are re-designed, which we will elaborate later.
% For instance, the algorithms to handle moving objects are different.
% Note that additional control points could be added to further improve alignment performance in the step 2).

\begin{figure}
  \centering
  \includegraphics[width=\linewidth]{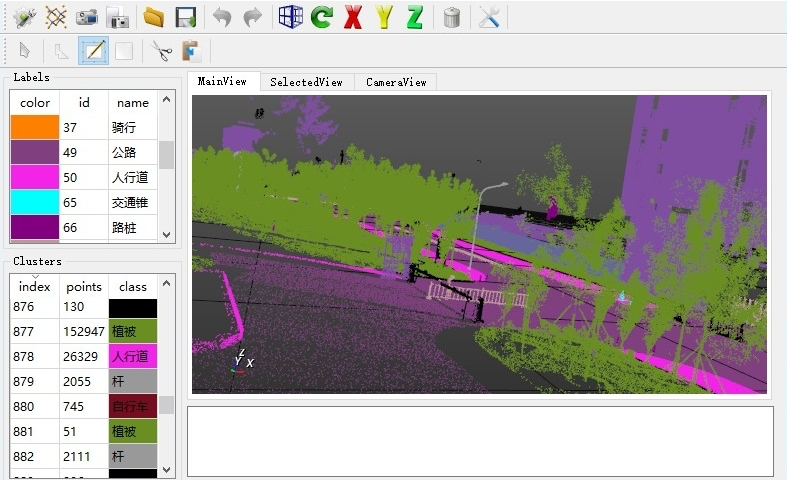}
  \caption{The user interface of our 3D labeling tool. At left-top, we show the pre-defined color code of different classes. At left-bottom, we show the labelling logs which can be used to revert the labelling when mistakes happen. At center part, labelled point cloud is shown indicating the labelling progress.}
  \label{fig:tool}
  \vspace{-0.5\baselineskip}
\end{figure}

\begin{figure*}[t]
\begin{center}
\includegraphics[width=\linewidth]{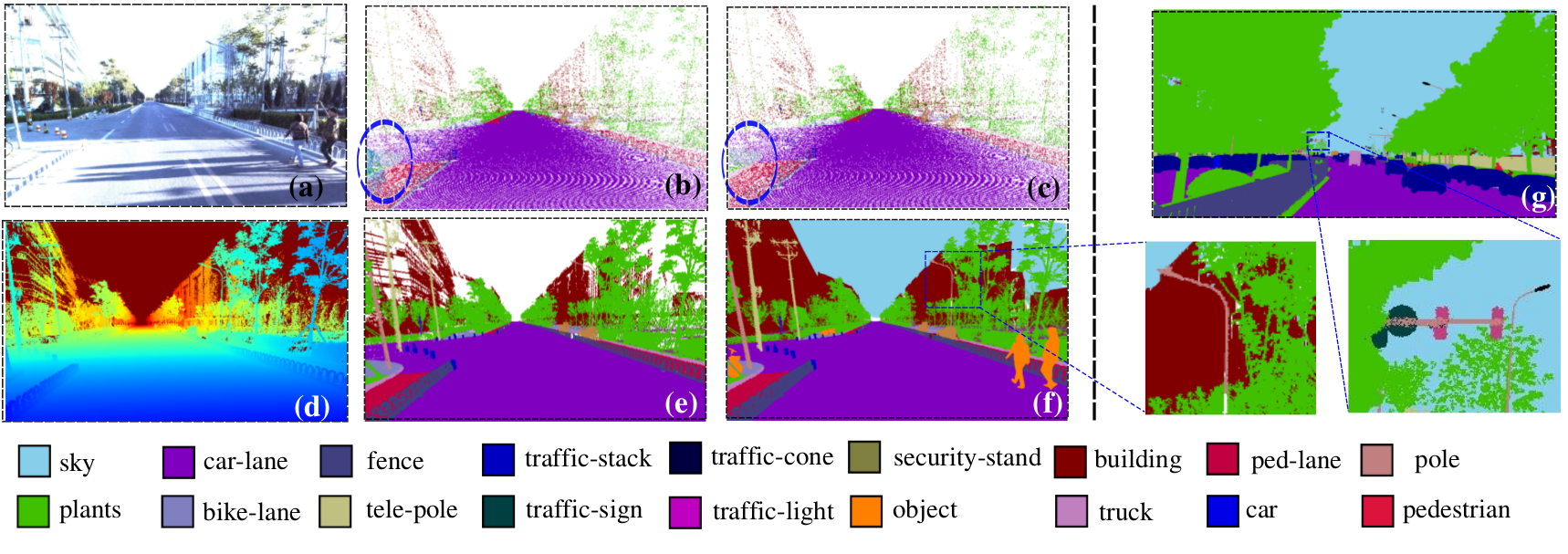}
\end{center}
	\caption{A labelled example of the labelling pipeline for semantic parsing, a subset of color coded labels are shown below. (a) Image. (b) Rendered label map with 3D point cloud projection, with an inaccurate moving object (rider) circled in blue. (c) Rendered label map with 3D point cloud projection after points with low temporal consistency being removed. (d) $\&$ (e) Rendered depth map of background and rendered label map after class dependent splatting in 3D point clouds (\secref{subsec:label}). (f) Merged label map with missing region in-painted, moving objects and sky. (g) Another label map with very small traffic lights. Details are zoomed out highlighting the details of our rendered label maps. Other examples of our labeled videos is shown online~\cite{segment}.}
\label{fig:data}
\vspace{-1.3\baselineskip}
\end{figure*}

\medskip
\noindent\textbf{Moving object removal.} As mentioned in Sec.~\ref{subsec:acq}, LIDAR scanner $Riegl$ is accurate in static background, while due to low scanning rate, the point clouds of moving objects, such as vehicles and pedestrians running on the road, could be compressed, expanded, or completely missing in the captured point clouds as illustrated in Fig.~\ref{fig:data}(b). Thus, we design to handle labelling static background and moving object separately, as shown in Fig.~\ref{fig:labeling}. Specifically, in the first step, we do moving object removal from our collected point clouds by 1) scan the same road segment multiple rounds; 2) align these point clouds based on manually selected control points; 3) remove the points based on the temporal consistency. Formally, the condition to kept a point $\ve{x}$ in round $j$ is,
% {\vspace{-0.5\baselineskip}
\begin{align}
\sum_{i=0}^{r}{\mathbbm{1}(\exists~\ve{x}_i \in \hua{P}_i~s.t.~\|\ve{x}_i - \ve{x}_j\| < \epsilon_d )} / r \geq \delta
\end{align}
% }
where $\delta = 0.6$ and $\epsilon_d = 0.025m$ in our setting, and $\mathbbm{1}()$ is an indicator function. It indicates that a 3D point will be kept if it appears with high frequency in many rounds of recording, \ie 60$\%$ of all times.
We keep the remained point clouds as a static background $\hua{M}$ for semantic labelling.
\medskip
\noindent\textbf{3D labelling.}
Next, for labelling static background (3D Labeling), rather than label each 3D point and loading all the points, we first separate the 3D points into multiple parts, and over-segment each part of point clouds into point clusters based on spatial distances and normal directions using locally convex connected patches (LCCP)~\cite{christoph2014object} implemented with PCL~\cite{pcl}.
Then, we label these point clusters manually using our in-house developed 3D labelling tool as shown in Fig.~\ref{fig:tool}, which can easily do point cloud rotation, (inverse-)selection by polygons, matching between point clouds and camera views, etc.. %and we will release this tool later jointly with this paper.
Notice at this stage, there will be point clouds belonging to movable but static objects such as bicycles and cars parking aside the road. These point clouds are remained in our background, and also labelled in 3D which are valuable to increase our label efficiency of objects in 2D images. % Thus, we also label these point cloud

To further improve 3D point cloud labelling efficiency, after labelling of one road, we actively train a PointNet++ model~\cite{qi2017pointnet++} to pre-label the over-segmented point cloud clusters of the next road. Labellers are then asked to refine and correct the results by fixing wrong annotations, which often occur near the object boundaries. With the growing number of labelled point clouds, our learned model can label new roads with increasing accuracy, yielding accelerated labelling process, which scales up to various cities and roads.  % Our 3D active labeling tool integrates all these modules together, yielding speed up the labeling process, which includes
% The user interface design of the tool as shown in Figure~\ref{fig:tool}

\medskip
\noindent\textbf{Splatting $\&$ Projection.}
Once the 3D annotations are generated, the annotations of static background/objects for all the 2D image frames are generated automatically by 3D-2D projections. In our setting, the 3D map is a point cloud based environment.
Although the density of the point cloud is very high (one point per 25mm within road regions), when the 3D points are far away from the camera, the projected labels could be sparse, \eg regions of buildings shown in \figref{fig:data}(c).
Thus for each point in the environment, we adopt the point splatting technique, by enlarging the 3D point to a square where the square size is determined by its semantic class.

Formally, given a 6-DOF camera pose $\ve{p} = [\ve{q}, \ve{t}] \in SE(3)$, where $\ve{q} \in SO(3)$ is the quaternion representation of rotation and $\ve{t} \in \mathbbm{R}^3$ is translation, a label map can be rendered from the semantic 3D map, where z-buffer is applied to find the closest point at each pixel. For a 3D point $\ve{x}$ belonging a class $c$, its square size $s_c$ is set to be proportional to the class' average distance to the camera. Formally,
{\vspace{-0.3\baselineskip}
\begin{align}
\label{eq:square_size}
s_c \propto \frac{1}{|\hua{P}_c|}\sum_{\ve{x}\in \hua{P}_c} \min_{\ve{t}\in\hua{T}} d(\ve{x}, \ve{t})
\end{align}
}
where $\hua{P}_c$ is the set of 3D points belong to class $c$, and $\hua{T}$ is the set of ground truth camera poses.  Then, given the relative square size between different classes, we define an absolute range to obtain the actual square size for splatting. This is non-trivial since too large size will result in dilated edges, while too small size will yield many holes. In our experiments, we set the range as $[0.025, 0.05]$, and find that it provides the highest visual quality.
As shown in \figref{fig:data}(e), invalid values in-between those projected points are well in-painted, meanwhile the boundaries separating different semantic classes are also well preserved, yielding the both the background depth map and 2D labelled background.
With such a strategy, we increase labelling efficiency and accuracy for video frames. For example, it could be very labor-intensive to label texture-rich regions like trees, poles and traffic lights further away, especially when occlusion happens like fence on the road as illustrated in Fig.~\ref{fig:data}(g).

\medskip
\noindent\textbf{2D labelling of objects and backgrounds.}
Finally, to generate the final labels (\figref{fig:data}(f)), we need to label the moving objects in the environments, and fix missing parts at background like part of building regions.
Similar with 3D point cloud labelling, we also developed an in-house 2D labelling tool with the same interface as 3D tool in Fig.~\ref{fig:tool}.
To speed up the 2D semantic labeling, we also use a labelling strategy by training a CNN network for movable objects and background~\cite{wu2016wider} to pre-segment the 2D images. For segmenting background, we test with original image resolution collected by our camera, where the resolution is much higher than that used in the original paper to increase the quality of predicted region boundaries. For segment objects, similar with MaskRCNN~\cite{he2018mask}, we first do 2D object detection with faster RCNN~\cite{ren2015faster}, and segment object masks inside.
However, since we consider high requirements for object boundaries rather than class accuracy, for each bounding box with high confidence ($\geq 0.9$), we enlarge the bounding box and crop out the object region with context similar to~\cite{wang2015joint}. Then, we upsample the cropped image to a higher resolution by setting a minimum resolution of prediction (minimum len greater than $512$), and segment out the mask with an actively trained mask CNN network with the same architecture in~\cite{wu2016wider}. The two networks for segmenting background and objects are updated when images in one road is labelled. Here, the learning parameters from these networks follow the original papers.

Finally, the segmented results from the networks are fused with our rendered label map from the semantic 3D point clouds following two rules: 1) for fusing segmented label map from the background network, we fill the predicted label in the pixels without 3D projection, yielding a background semantic map. 2) for fusing semantic object label segmented by object network, we pasted the object mask over the fused background map, without replacing the projected static movable object mask rendered from 3D points as mentioned in 3D labelling. We provided this fused label map for labellers to further fine tuning when error happens especially around object boundary or occlusion from the object masks. In addition, the user can omit any of the pre-segmented results from CNNs to do relabelling if the segmented results are far from satisfaction. Our label tool supports multiple actions such as polygons and pasting brushes etc., which are commonly adopted by many popular open source label tools~\footnote{\url{https://github.com/topics/labeling-tool}}.

A final labelled example is shown in Fig.~\ref{fig:data}(f)$\&$(g). Notice that some background classes such as fence, traffic light, and vegetation are annotated in details using our projection and missing parts such as building glass can be fill in. Thanks to 3D and active learning, our overall pipeline save us significant efforts in dense per-pixel and per-frame semantic labelling for background and objects.
In practice, our labelling pipeline can reduce the time cost of dense labelling task per-image from nearly 1 hour to around 10 minutes, with the guarantee of passing our quality control process.

% In other datasets, these classes could be ambiguous caused by occlusions or labeled as a whole region in order to save labeling efforts.

% Another labeling tool for 2D images is developed to fix or refine the segmentation results.
% Again, the wrong annotations often occur around the object boundaries that could be caused by merge/split of multiple objects and harsh lighting conditions.
% Our 2D labeling tool is designed so that the control points of the boundaries could be easily selected and adjusted.

% \begin{figure}[t]
%   \centering
%   \includegraphics[width=\linewidth]{./fig/label_exp.pdf}
%   \caption{An example of 2D annotation with boundaries in details. Color inconsistent with Fig.~\ref{fig:data} due to different visualization tools.}\label{fig:label_exp}
% \end{figure}
\begin{figure*}[t]
\begin{center}
\includegraphics[width=0.8\linewidth]{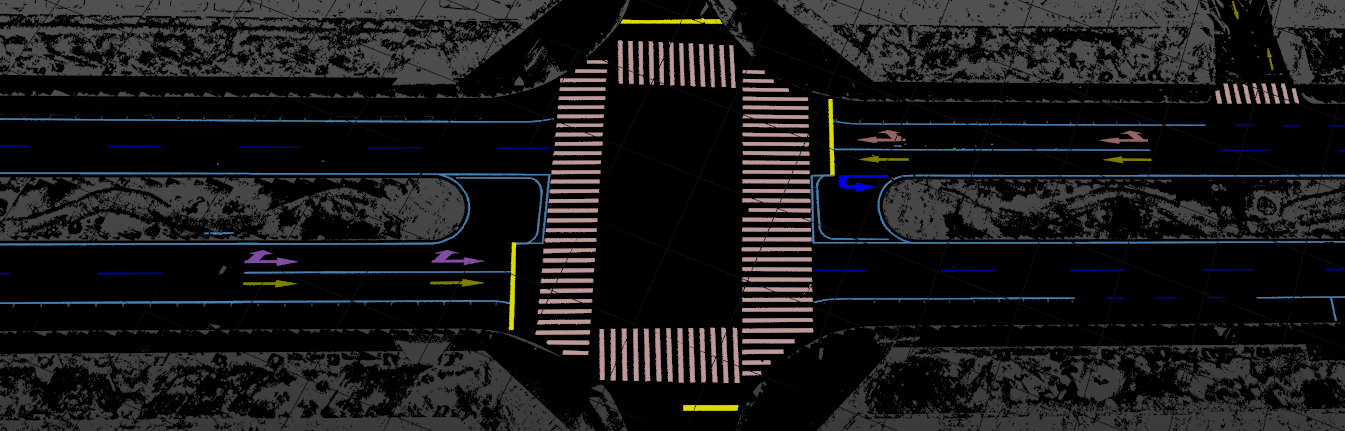}
\end{center}
	\caption{Bird view of our projected road lane marks with labelling.}
\label{fig:lane}
\vspace{-1.3\baselineskip}
\end{figure*}

\medskip
\noindent\textbf{Labelling of lane mark segments on road.} In self-driving, lane marks are information solely from static background. Fortunately, our collected survey-grade 3D points not only have high density, but also contain lighting intensity, dependent on which we can distinguish the lane mark on the roads.
Specifically, we perform similar labelling process as 3D labelling of rigid background by labelling each 3D point  to pre-defined lane mark labels listed in Tab.~\ref{tb:class2}.

Nevertheless, different from labelling 3D point clusters where point clouds from buildings and trees are important, for lane marks, we only need to consider points on the road. Therefore, we take out the road point clouds based on normal directions, and perform orthogonal projection of these points from the bird view to a high resolution 2D image, as shown in Fig.~\ref{fig:lane}, over which labellers draw a polygon for each lane mark on the road. In the meantime, our tool brings out the corresponding images, and highlights the regions in 2D for each labelled polygon, where the color and type of the labelled lanemark can be determined.
%Here to make sure a 3D point cluster of lane mark is well segmented, we add point intensity into the over-segmentation metrics.
% In,  determining the color of the lane mark, our tool automatically brings out the corresponding images

\medskip
\noindent\textbf{Labelling of instance segments.} Thanks to an active labelling component with detection, it is easy for us to generalize the segmentation label map to produce instance masks given the segmented results from the object detection and segmentation networks.
Specifically, we ask the labellers to refine the boundary between different instances when it is necessary, \ie visually significantly not aligned with true object boundaries.

\medskip
\noindent\textbf{Control of label quality. } Following the existing standard work flows of crowdsourcing object annotations~\cite{su2012crowdsourcing,kovashka2016crowdsourcing,li2016crowdsourced}, all our 2D/3D labeling tasks, \eg 3D point cloud, 2D background, 2D instance and 3D lanemark, contain verification stages to control the label quality. %For instance, in our pipeline (\figref{fig:labeling}), we have three labelling tasks, \ie~3D point cloud labelling, 2D background labelling and 2D instance labelling.
Specifically, for each task, we have a detailed instruction to train our labellers, and a labeller is good to start labelling after passing a designed quiz. % \figref{fig:qa} illustrates two specific rules for the 2D instance annotation and lane annotation.
We will publish all our instructions on our website to benefit the community upon the publication of this paper.

After the labelling stage, we have a review stage, and each reviewer is an experienced labeller had sufficient labelled images (over 500) passed our label quality verification. The reviewer will verify the quality and the coverage of labelled regions. In addition, since we do video labelling, we also have reviewer to visually verify the semantics are temporally consistent in the next frame. Only if an image has passed two reviewers, it could be accepted as a valid ground truth.

% Another set of labellers then verify whether all the objects are correctly labeled in an image and corresponding labels are temporally consistent within $\pm3$ frames.

% In the drawing stage, we define precise annotation rules and make sure that the labellers understand them.
% This is done by a labeler training process based on instructions and qualification tests.
% Our rules are defined to cover all the specific cases and revised during the drawing stage.

% \begin{figure}
%   \centering
%   \includegraphics[width=\linewidth]{fig/qa.pdf}
%   \caption{Example labeling rules. (\textit{left}) Vehicles are disjointed by occlusion.  During the 2D instance labeling, disjointed vehicle parts should be labeled and assigned to the same instance ID. \textit{(right)} Lane markings are often invisible on city sewer covers. During the 3D labeling, the invisible lane markings on covers should be completed manually.}\label{fig:qa}
% \end{figure}
\medskip
\noindent\textbf{Existing issues.} LiDAR scanners could fail on translucent and highly reflective surfaces such as mirrors of buildings. Though we fixed this problem in part of our recorded videos, \eg as shown in \figref{fig:data}), we found it is still over laborious to fix every frame in all our videos even with active labelling. Therefore, part of video frames in our current release, pixels without 3D projection or active labelling, \eg~sky and part of building in \figref{fig:example}, are set as \textit{void}, so that they are ignored during the training and evaluation. We leave labelling of these pixels to our future work.

\section{Deep localization and segmentation}
\label{sec:localize_and_parsing}

\begin{figure*}[t]
\center
%\vspace{-1\baselineskip}
%\includegraphics[width=0.9\textwidth]{png/framework_150ppi.png}
\includegraphics[width=0.9\textwidth]{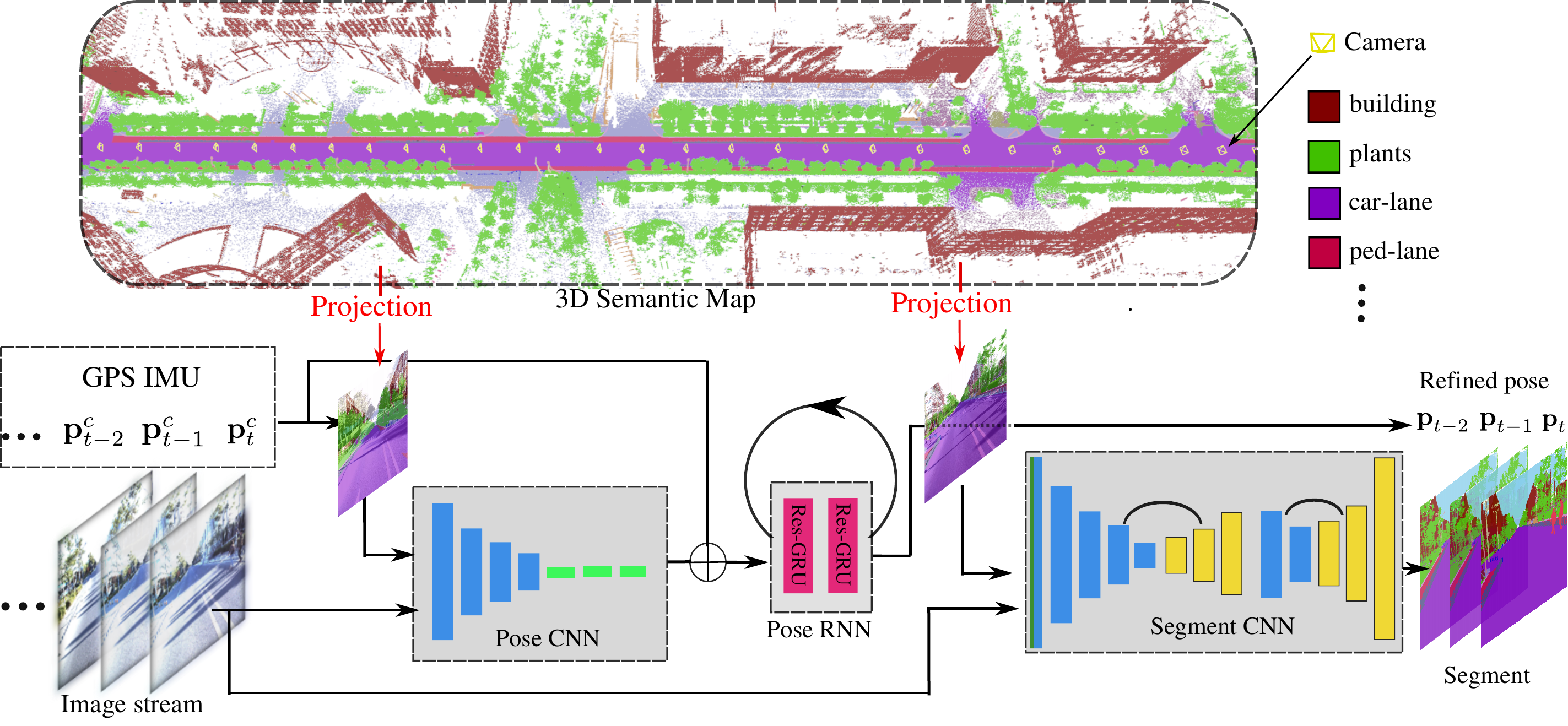}
\caption{DeLS-3D overview. The black arrows show the testing process, and red arrows indicate the rendering (projection) operation in training and inference. The yellow frustum shows the location of cameras inside the 3D map. The input of our system contains a sequence of images and corresponding GPS/IMU signals. The outputs are the semantically segmented images, each with its refined camera pose.}
\label{fig:framework}
\vspace{-1\baselineskip}
\end{figure*}

As discussed in introduction (Sec.~\ref{sec:introduction}), ApolloScape contains various ground truth which enables multitask learning. In this paper, we show such a case by creating a deep learning based system for joint localization and semantic segmentation given a semantic 3D map~\cite{wang2018dels}, which we call DeLS-3D, as illustrated in \figref{fig:framework}. Specifically, at upper part, a pre-built 3D semantic map is available. During testing, an online stream of images and corresponding coarse camera poses from GPS/IMU are fed into the system. Firstly, for each frame, a semantic label map is rendered out given the input coarse camera pose, which is fed into a pose CNN jointly with the respective RGB image. The network calculates the relative rotation and translation, and yields a corrected camera pose. To incorporate the temporal correlations, the corrected poses from pose CNN are fed into a pose RNN to further improves the estimation accuracy in the stream.
Last, given the rectified camera pose, a new label map is rendered out, which is fed together with the image to a segment CNN. The rendered label map helps to segment a spatially more accurate and temporally more consistent result for the image stream of video.
In this system, since \emph{ApolloScape} contains ground truth for both camera poses and segments, it can be trained with strong supervision at each end of outputs. The code for our system has been released at~\url{https://github.com/pengwangucla/DeLS-3D}. % the full system is based on a semantic 3D map and deep networks.
In the following, we elaborate our network architectures and the loss functions to train the whole system.

% By feeding in such a rendered map, it facilitates the CNN to better discover scene layouts both for pose estimation and scene parsing.
% Another possibility is to render a RGB image rather than a label map. However, we found label map works better, which may be due to it contains less noisy edges than rendered RGB image. \textcolor{red}{it is not certain, should we even bother to say it?}
% AOne may wonder why we render semantic label map rather than synthesizing a RGB image for later inputs.
% This is because intuitively label maps is invariant to color changes, which yields better boundaries reflecting the scene layout.

\subsection{Camera localization with motion prior}
\medskip
\noindent\textbf{Translation rectification with road prior.} One common localization priori for navigation is to use the 2D road map, by constraining the GPS signals inside the road regions. We adopt a similar strategy, since once the GPS signal is out of road regions, the rendered label map will be totally different from the street-view, and no correspondence can be found by the network.

To implement this constraint, firstly we render a 2D road map image with a rasterization grid of $0.05m$ from our 3D semantic map by using only road 3D points, \ie points belong to car-lane, pedestrian-lane and bike-lane \etc
Then, at each pixel $[x, y] \in \mathbbm{Z}^2$ in the 2D map, an offset value $\ve{f}(x, y)$ is pre-calculated indicating its 2D offset to the closest pixel belongs to road through the breath-first-search (BFS) algorithm efficiently.

During online testing, given a noisy translation $\ve{t}=[t_x, t_y, t_z]$, we can find the closest road points \wrt $\ve{t}$ using $[t_x, t_y] + \ve{f}(\lfloor t_x \rfloor, \lfloor t_y \rfloor)$ from our pre-calculated offset function. Then, a label map is rendered based on the rectified camera pose, which is fed to pose CNN.

\begin{figure}[b]
\begin{center}
\includegraphics[width=.8\linewidth]{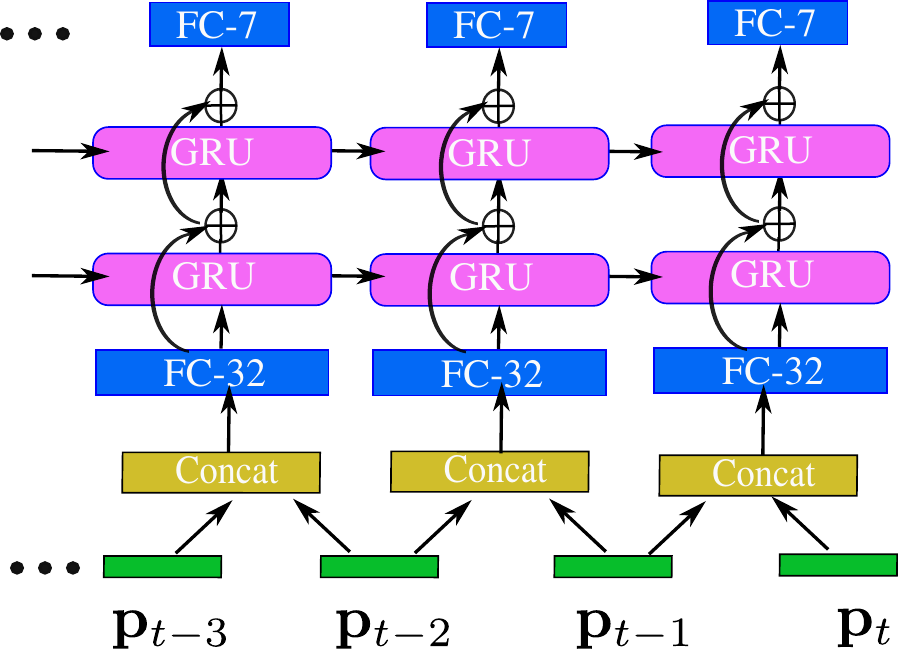}
\end{center}
   \caption{The GRU RNN network architecture for modeling a sequence of camera poses.}
\label{fig:rnn}
\vspace{-1.3\baselineskip}
\end{figure}
\medskip
\noindent\textbf{CNN-GRU pose network architecture.}
As shown in \figref{fig:framework}, our pose networks contain a pose CNN and a pose GRU-RNN. Particularly,
the CNN of our pose network takes as inputs an image $\ve{I}$ and the rendered label map $\ve{L}$ from corresponding coarse camera pose $\ve{p}_i^c$. It outputs a 7 dimension vector $\hat{\ve{p}}_i$ representing the relative pose between the image and rendered label map, and we can get a corrected pose \wrt the 3D map by $\ve{p}_i = \ve{p}_i^c + \hat{\ve{p}_i}$.
For the network architecture of pose CNN, we follow the design of DeMoN~\cite{ummenhofer2016demon}, which has large kernel to obtain bigger context while keeping the amount of parameters and runtime manageable. The convolutional kernel of this network consists a pair of 1D filters in $y$ and $x$-direction, and the encoder gradually reduces the spatial resolution with stride of 2 while increasing the number of channels. We list the details of the network in our implementation details at \secref{sec:experiments}.

Additionally, since the input is a stream of images, in order to model the temporal dependency,
after the pose CNN, a multi-layer GRU with residual connection~\cite{wu2016google} is appended.
More specifically, we adopt a two layer GRU with 32 hidden states as illustrated in \figref{fig:rnn}. It includes high order interaction beyond nearby frames, which is preferred for improve the pose estimation performance.
In traditional navigation applications of estimating 2D poses,  Kalman filter~\cite{kalman1960new} is commonly applied by assuming either a constant velocity or acceleration.
In our case, because the vehicle velocity is unknown, transition of camera poses is learned from the training sequences, and in our experiments we show that the motion predicted from RNN is better than using a Kalman filter with a constant speed assumption, yielding further improvement over the estimated ones from our pose CNN.

\begin{figure*}[t]
\center
%\vspace{-0.6\baselineskip}
%\includegraphics[width=0.95\textwidth]{png/segCNN_600ppi.png}
\includegraphics[width=0.95\textwidth]{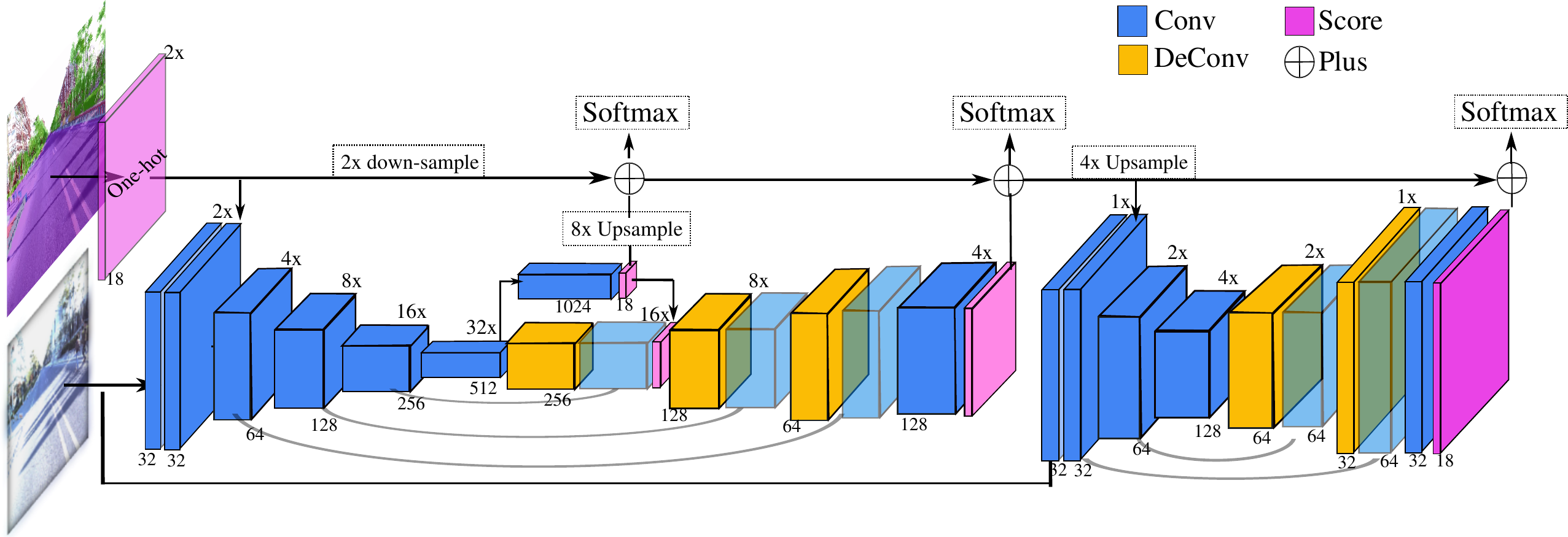}
\caption{Architecture of the segment CNN with rendered label map as a segmentation priori. At bottom of each convolutional block, we show the filter size, and at top we mark the downsample rates of each block \wrt the input image size. The `softmax' text box indicates the places a loss is calculated. Details are in \secref{subsec:parsing}.}
\label{fig:segnet}
\vspace{-1.35\baselineskip}
\end{figure*}

\medskip
\noindent\textbf{Pose loss.}
Following the PoseNet~\cite{kendall2017geometric}, we use the geometric matching loss for training, which avoids the balancing factor between rotation and translation.
Formally, given a set of point cloud in 3D $~\hua{P}=\{\ve{x}\}$, and the loss for each image is written as,
{
\begin{align}
L(\ve{p}, \ve{p}^*) = \sum_{\ve{x} \in \hua{P}}\omega_{l_\ve{x}}|\pi(\ve{x}, \ve{p}) - \pi(\ve{x}, \ve{p}^*)|_2
\label{eq:proj_loss}
\end{align}
}
where $\ve{p}$ and $\ve{p}^*$ are the estimated pose and ground truth pose respectively. $\pi()$ is a projective function that maps a 3D point $\ve{x}$ to 2D image coordinates. $l_\ve{x}$ is the semantic label of $\ve{x}$ and $\omega_{l_\ve{x}}$ is a weight factor dependent on the semantics. Here, we set stronger weights for point cloud belong to certain classes like traffic light, and find it helps pose CNN to achieve better performance.
In~\cite{kendall2017geometric}, only the 3D points visible to the current camera are applied to compute this loss to help the stability of training. However, the amount of visible 3D points is still too large in practical for us to apply the loss.
Thus, we pre-render a depth map for each training image with a resolution of $256 \times 304$ using the ground truth camera pose, and use the back projected 3D points from the depth map for training.

% Intuitively, we find the amount of points in each class is dramatically unbalanced, and most points are those on roads and trees. Nevertheless, the appearance variation of road and trees are not very sensitive to pose changes in the street-view scenario. In order to encourage the network to discover more str
% Thus,  other structures images like electricity pole or road light have rich structures like edges and textures, which potentially should be valued more for matching.
% In~\cite{ummenhofer2016demon}, the model try to predict a flow confidence map revealing the texture rich regions which helps pose estimation. In our work, we adopt the labelled semantic labels, and reweight each 3D point in the training loss, which drives the network to focus more on
% Thus, the learning loss changed to
% \begin{align}
% L(\ve{p}, \ve{p}^*) = \sum_{\ve{x} \in \hua{P}}\omega_{l_\ve{x}}|\pi(\ve{x}, \ve{p}) - \pi(\ve{x}, \ve{p}^*)|_2
% \label{eq:proj_loss}
% \end{align}
% where $\omega_{l_\ve{x}}$ is the weight of class $l_\ve{x}$. We set the weight for each class depends on the class edgeness which is the percentage of pixel along the edge

\subsection{Video parsing with pose guidance}
\label{subsec:parsing}

Having rectified pose at hand, one may direct render the semantic 3D world to the view of a camera, yielding a semantic parsing of the current image. However, the estimated pose is not perfect, fine regions such as light poles can be completely misaligned. Other issues also exist. For instance, many 3D points are missing due to reflection, \eg regions of glasses, and points can be sparse at long distance. Last, dynamic objects in the input cannot be represented by the projected label map, yielding incorrect labelling at corresponding regions. Thus, we propose an additional segment CNN to tackle these issues, while taking the rendered label map as segmentation guidance.
% In our experiments, we show with pose rendered label map as an additional input, the segment results are temporally more consistent and yields better accuracy.

\medskip
\noindent\textbf{Segment network architecture.} As discussed in \secref{sec:related}, heavily parameterized networks such as ResNet are not efficient enough for our online application. Thus, as illustrated in \figref{fig:segnet}, our segment CNN is a light-weight network containing an encoder-decoder network and a refinement network, and both have similar architecture with the corresponding ones used in DeMoN~\cite{ummenhofer2016demon} including 1D filters and mirror connections. However, since we have a segment guidance from the 3D semantic map, we add a residual stream (top part of \figref{fig:segnet}), which encourages the network to learn the differences between the rendered label map and the ground truth. In~\cite{pohlen2016full}, a full resolution stream is used to keep spatial details, while here, we use the rendered label map to keep the semantic spatial layout.

Another notable difference for encoder-decoder network from DeMoN is that for network inputs, shown in \figref{fig:segnet}, rather than directly concatenate the label map with input image, we transform the label map to a score map through one-hot operation, and embed the score of each pixel to a 32 dimensional feature vector.
% This is to balance the two.\textcolor{red}{not very clear here, how is this balanced?}
Then, we concatenate this feature vector with the first layer output from image, where the input channel imbalance between image and label map is alleviated, which is shown to be useful by previous works~\cite{eigen2015predicting}.
For refinement network shown in \figref{fig:segnet}, we use the same strategy to handle the two inputs.
Finally, the segment network produces a score map, yielding the semantic parsing of the given image.

We train the segment network first with only RGB images, then fine-tune the network by adding the input of rendered label maps. This is because our network is trained from scratch, therefore it needs a large amount of data to learn effective features from images. However, the rendered label map from the estimated pose has on average 70$\%$ pixel accuracy, leaving only 30$\%$ of pixels having effective gradients. This could easily drive the network to over fit to the rendered label map, while slowing down the process towards learning features from images. Finally, for segmentation loss, we use the standard softmax loss, and add intermediate supervision right after the outputs from both the encoder and the decoder as indicated in \figref{fig:segnet}.

\section{Experiments} \label{sec:experiments}
In this section, we first evaluate our online deep localization and segmentation algorithms (DeLS-3D) on two of our released roads, which is a subset of our full data. We compare it against other SOTA deep learning based visual localization, \ie PoseNet~\cite{Kendall_2015_ICCV}, and segmentation algorithms \ie ResNet38~\cite{wu2016wider}, which shows the benefits of multitask unification.

Then, we elaborate the benchmarks setup online with ApolloScape and the current leading results, which follows many standard settings such as the ones from KITTI~\cite{Geiger2013IJRR} and Cityscapes~\cite{Cordts2016Cityscapes}. These tasks include semantic segmentation, semantic instance segmentation, self-localization, lanemark segmentation. Due to the ``DeLS'' algorithm proposed in this work does not follow those standard experimental settings, we could not provide its results for the benchmarks. Nevertheless, for each benchmark, we either ran a baseline result with SOTA methods or launched a challenge for other researchers, providing a reasonable estimation of the task difficulties.

\subsection{Evaluate DeLS-3D}
\label{subsec:dels-3d}

In this section, we evaluate various settings for pose estimation and segmentation to validate each component in the DeLS-3D system.
For GPS and IMU signal, despite we have multiple scans for the same road segments, it is still very limited for training. Thus, follow~\cite{vishal2015accurate}, we simulate noisy GPS and IMU by adding random perturbation $\epsilon$ \wrt the ground truth pose following uniform distributions.
Specifically, translation and rotation noise are set as $\epsilon_t \sim U(0, 7.5m)$ and $\epsilon_r \sim U(0^{\circ}, 15^{\circ})$ respectively.
We refer to realistic data~\cite{lee2015gps} for setting the noisy range of simulation.

\medskip
\noindent\textbf{Datasets.} Two roads early collected at Beijing in China are used in our evaluation.
The first one is inside a technology park, named zhongguancun park (Zpark), and we scanned 6 rounds during different daytimes. The 3D map generated has a road length around 3$km$, and the distance between consecutive frames is around 5$m$ to 10$m$. We use 4 rounds of the video camera images for training and 2 for testing, yielding 2242 training images and 756 testing images.
The second one we scanned 10 rounds and 4km near a lake, named daoxianghu lake (Dlake), and the distance between consecutive frames is around 1$m$ to 3$m$.
We use 8 rounds of the video camera images for training and 2 for testing, yielding 17062 training images and 1973 testing images. The existing semantic classes in the two datasets are shown in \tabref{tbl:segment}, which are subsets from our full semantic classes.
% We will release the two datasets separately. %include $\{$sky, car-lane, pedestrian-lane, bike-lane, curb, traffic-cone, traffic-stack, traffic-fence, light-pole, traffic-light, tele-pole, traffic-sign, billboard, building, security-stand, plants, car, $\}$.

\medskip
\noindent\textbf{Implementation details.} To quickly render from the 3D map, we adopt OpenGL to efficiently render a label map with the z-buffer handling. A 512 $\times$ 608 image can be generated in 70ms with a single Titan Z GPU, which is also the input size for both pose CNN and segment CNN.
For pose CNN, the filter sizes of all layers are $\{32, 32, 64, 128, 256, 1024, 128, 7\}$, and the forward speed for each frame is 9ms. For pose RNN, we sample sequences with length of 100 from our data for training, and the speed for each frame is 0.9ms on average.
For segment CNN, we keep the size the same as input, and the forward time is 90ms. Overall, the inference speed is around 240ms per-image for performing joint localization and segmentation.
Both of the network is learned with 'Nadam' optimizer~\cite{dozat2016incorporating} with a learning rate of $10^{-3}$. We sequentially train these three models due to GPU memory limitation.
Specifically, for pose CNN and segment CNN, we stops at 150 epochs when there is no performance gain, and for pose RNN, we stops at 200 epochs. For data augmentation, we use the imgaug\footnote{\url{https://github.com/aleju/imgaug}} library to add lighting, blurring and flipping variations. We keep a subset from training images for validating the trained model from each epoch, and choose the model performing best for evaluation.

For testing, since input GPS/IMU varies every time, \ie~$\ve{p}_t^c=\ve{p}^*+\epsilon$, we need to have a confidence range of prediction for both camera pose and image segment, in order to verify the improvement of each component we have is significant. Specifically, we report the standard variation of the results from a 10 time simulation to obtain the confidence range. Finally, we implement all the networks by adopting the MXNet~\cite{ChenLLLWWXXZZ15} platform.

\medskip
\noindent\textbf{Evaluation metrics.} We use the median translation offset and median relative angle~\cite{Kendall_2015_ICCV}. For evaluating segment, we adopt the commonly used pixel accuracy (Pix. Acc.), mean class accuracy (mAcc.) and mean intersect-over-union (mIOU) as that from~\cite{wu2016wider}.

\begin{table}
\center
\fontsize{6.5}{7}\selectfont
%\hspace*{-0.23cm}
\begin{tabular}{lllll}
\toprule[0.1 em]
% \thickhline
%& PoseNet~\cite{kendall2017geometric} & -  & -  & -  \\
Data & Method & Trans (m) $\downarrow$ & Rot ($\circ$)$\downarrow$  & Pix. Acc($\%$)$\uparrow$ \\
\hline
\parbox[t]{2mm}{\multirow{7}{*}{\rotatebox[origin=c]{90}{Zpark}}} & Noisy pose & 3.45 $\pm$ 0.176 & 7.87 $\pm$ 1.10 & 54.01 $\pm$ 1.5 \\
& Pose CNN w/o semantic & 1.355 $\pm$ 0.052  & 0.982 $\pm$ 0.023 & 70.99 $\pm$ 0.18 \\
& Pose CNN w semantic & 1.331 $\pm$ 0.057  & 0.727 $\pm$ 0.018 & 71.73 $\pm$ 0.18  \\
& Noisy pose w KF & 2.56 $\pm$ 0.16 & 7.37 $\pm$ 1.01 & 55.1 $\pm$ 0.91 \\
& Pose RNN w/o CNN & 1.282 $\pm$ 0.061  & 1.731 $\pm$ 0.06 &  68.10 $\pm$ 0.32 \\
& Pose CNN w KF & 1.281 $\pm$ 0.06  & 0.833 $\pm$ 0.03 & 72.00 $\pm$ 0.17  \\
& Pose CNN-RNN  & \textbf{1.005} $\pm$ 0.044  & \textbf{0.719} $\pm$ 0.035  & \textbf{73.01} $\pm$ 0.16  \\
\toprule[0.1 em]
\hline
\parbox[t]{1mm}{\multirow{3}{*}{\rotatebox[origin=c]{90}{Dlake}}} & Pose CNN w semantic & 1.667 $\pm$ 0.05 & 0.702 $\pm$ 0.015 & 87.83 $\pm$ 0.017 \\
& Pose RNN w/o CNN & 1.385 $\pm$ 0.057 & 1.222 $\pm$ 0.054 & 85.10 $\pm$ 0.03\\
& Pose CNN-RNN  & \textbf{0.890} $\pm$ 0.037  & \textbf{0.557}$\pm$ 0.021 & \textbf{88.55} $\pm$ 0.13  \\
\toprule[0.1 em]
\end{tabular}
\caption{Compare the accuracy of different settings for pose estimation from the two datasets.
Noisy pose indicates the noisy input signal from GPS, IMU, and 'KF' means kalman filter.
The number after $\pm$ indicates the standard deviation (S.D.) from 10 simulations. $\downarrow \& \uparrow$ means lower the better and higher the better respectively.  We can see the improvement is statistically significant.}
\label{tbl:pose}
\vspace{-1.5\baselineskip}
\end{table}

\begin{table*}[t]
\center
%\vspace{-0.5\baselineskip}
\fontsize{7}{7}\selectfont
\setlength\tabcolsep{1.5pt}

\begin{tabular}{llcccccccccccccccccccc}
%\vspace{-1.0\baselineskip}
% \toprule[0.1 em]
%\thickhline
Data & \multicolumn{1}{c}{Method} & \rot{mIOU}  & \rot{Pix. Acc}   & \rot{sky} & \rot{car-lane} & \rot{ped-lane} & \rot{bike-lane} & \rot{curb} & \rot{$t$-cone} & \rot{$t$-stack} & \rot{$t$-fence} & \rot{light-pole} & \rot{$t$-light} & \rot{tele-pole} & \rot{$t$-sign} & \rot{billboard} & \rot{temp-build} & \rot{building} & \rot{sec.-stand} & \rot{plants} & \rot{object} \\
\hline
\parbox[t]{2mm}{\multirow{6}{*}{\rotatebox[origin=c]{90}{Zpark}}} &
ResNet38~\cite{wu2016wider}     &64.66   & 95.87 &93.6 &98.5 &82.9 &87.2 &61.8 &46.1 &41.7 &82.0 &37.5 &26.7 &45.9 &49.5 &60.0 &85.1 &67.3 &38.0 &89.2 &66.3\\
&SegCNN w/o Pose             &68.35   & 95.61 &94.2 &98.6 &83.8 &89.5 &69.3 &47.5 &52.9 &83.9 &52.2 &43.5 &46.3 &52.9 &66.9 &87.0 &69.2 &40.0 &88.6 &63.8 \\
&Render PoseRNN              &32.61  & 73.1 & - & 91.7 &50.4 &62.1 &16.9 &6.6 &5.8 &30.5 &8.9 &6.7 &10.1 &16.3 &22.2 &70.6 &29.4 &20.2 &73.5 & - \\
&SegCNN w pose GT            &79.37   & 97.1  &96.1 &99.4 &92.5 &93.9 &81.4 &68.8 &71.4 &90.8 &71.7 &64.2 &69.1 &72.2 &83.7 &91.3 &76.2 &58.9 &91.6 &56.7 \\
&SegCNN w Pose CNN           &68.6  & 95.67  &94.5 &98.7 &84.3 &89.3 &69.0 &46.8 &52.9 &84.9 &53.7 &39.5 &48.8 &50.4 &67.9 &87.5 &69.9 &42.8 &88.5 &60.9 \\
&SegCNN w Pose RNN &\textbf{69.93}  &\textbf{95.98} &
                                             94.9 &98.8 &85.3 &90.2 &71.9 &45.7 &57.0 &85.9 &58.5 &41.8 &51.0 &52.2 &69.4 &88.5 &70.9 &48.0 &89.3 &59.5 \\
\toprule[0.1 em]
\end{tabular}

\begin{tabular}{llcccccccccccccccccccccc}
%\vspace{-2.0\baselineskip}
Data& \multicolumn{1}{c}{Method} & \rotf{mIOU} & \rotf{Pix. Acc}  & \rotf{sky} & \rotf{car-lane} & \rotf{ped-lane}  & \rotf{$t$-stack} & \rotf{$t$-fence} & \rotf{wall} & \rotf{light-pole} & \rotf{$t$-light} & \rotf{tele-pole} & \rotf{$t$-sign} & \rotf{billboard} & \rotf{building} & \rotf{plants} & \rotf{car} & \rotf{cyclist} & \rotf{motorbike} & \rotf{truck} & \rotf{bus}\\
\hline
\parbox[t]{2mm}{\multirow{3}{*}{\rotatebox[origin=c]{90}{Dlake}}} &
SegCNN w/o Pose  &62.36 &96.7 &95.3 &96.8 &12.8 &21.5 &81.9 &53.0 &44.7 &65.8 &52.1 &87.2 &55.5 &66.8 &94.5 &84.9 &20.3 &28.9 &78.4 &82.1 \\
&SegCNN w pose GT &73.10 &97.7 &96.8 &97.5 &41.3 &54.6 &87.5 &70.5 &63.4 &77.6 &70.5 &92.1 &69.2 &77.4 &96.1 &87.4 &24.5 &43.8 &80.0 &85.7 \\
&SegCNN w pose RNN &\textbf{67.00} &\textbf{97.1} &95.8 &97.2 &30.0 &37.4 &84.2 &62.6 &47.4 &65.5 &62.9 &89.6 &59.0 &70.3 &95.2 &86.8 &23.9 &34.4 &76.8 &86.6 \\
\toprule[0.1 em]
\end{tabular}
\caption{Compare the accuracy of different segment networks setting over Zpark (top) and Dlake (bottom) dataset. $t$ is short for 'traffic' in the table. Here we drop the 10 times S.D. to save space because it is relatively small ($\leq 0.1$). Our results are especially good at parsing of detailed structures and scene layouts, which is visualized in~\figref{fig:results}.}
\label{tbl:segment}
\vspace{-1.0\baselineskip}
\end{table*}

\medskip
\noindent\textbf{Pose Evaluation.}
We first directly compare with the work of PoseNet~\cite{Kendall_2015_ICCV,kendall2017geometric}, and use their published code and geometric loss (\equref{eq:proj_loss}) to train a model on Zpark dataset. Due to scene appearance similarity of the street-view, we did not obtain a reasonable model with their methods~\cite{Kendall_2015_ICCV,kendall2017geometric}, \ie results better than the noisy GPS/IMU signal.
\peng{Then, we experimented a leading open-source monocular SLAM algorithm, \ie ORB-SLAM~\cite{murTRO2015}, to do self-localization. However, it also provides no better result than provided initial poses, since low-level ORB features fail to match robustly due to many non-diffusion/reflective components in Zpark, such as glass buildings and specular new roads, plus repetitive appearance on trees.}
Therefore, in \tabref{tbl:pose}, we majorly list the performance of estimated translation $\ve{t}$ and rotation $\ve{r}$ from our model variations.
At the 1st row, we show the median error of GPS and IMU from our simulation.
At the 2nd row, by using our pose CNN with an additional input of projected label map, the model can learn good relative pose between camera and GPS/IMU, which significantly reduces the error (60$\%$ for $\ve{t}$, 85$\%$ for $\ve{r})$.
By adding semantic cues, \ie road priori and semantic weights in \equref{eq:proj_loss}, the pose errors are further reduced, especially for rotation (from $0.982$ to $0.727$ at the 3rd row). In fact, we found the most improvement is from semantic weighting, while the road priori helps marginally. In our future work, we would like to experiment larger noise and more data variations, which will better validate different cues.

\begin{figure*}[!htbp]
\center
%\vspace{-0.3\baselineskip}
\includegraphics[width=0.99\textwidth]{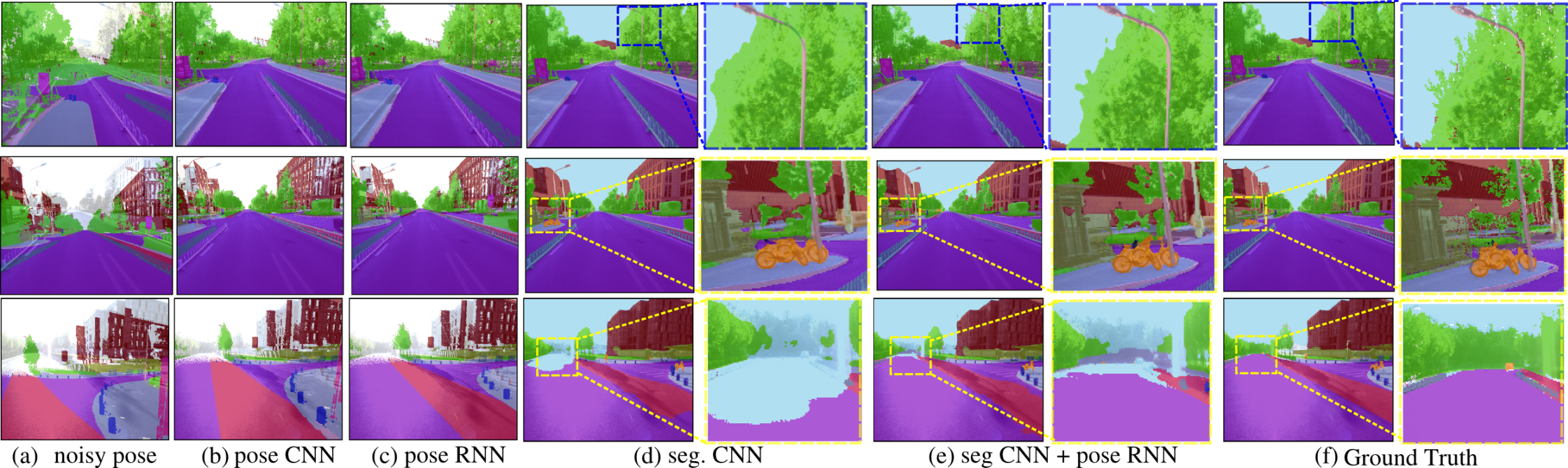}
   \caption{Results from each intermediate stage out of the system over Zpark dataset. Label map is overlaid with the image. Improved regions are boxed and zoomed out (best in color). More results are shown in the online videos for \href{https://youtu.be/HNPQVtgpjbE}{Zpark} and \href{https://youtu.be/ApyqPnvmJAs}{Dlake}.}
\label{fig:results}
\vspace{-0.6\baselineskip}
\end{figure*}

When having an video input, we first evaluate a simple baseline which refines the GPS/IMU  with Kalman filter~\cite{kalman1960new}, \ie~'Noisy pose w KF', which reasonably reduces the errors.
Then, we setup a baseline of performing RNN directly on the GPS/IMU signal, and as shown at 'Pose RNN w/o CNN', the estimated $\ve{t}$ is even better than pose CNN, while $\ve{r}$ is comparably much worse. This meets our expectation since the speed of camera is easier to capture temporally than rotation. Another baseline we adopt is performing  to the output from Pose CNN by assuming a constant speed which we set as the averaged speed from training sequences. As shown at 'Pose CNN w KF', it does improve slightly for translation, but harms rotation, which means the filter over smoothed the sequence. Finally when combining pose CNN and RNN, it achieves the best pose estimation both for $\ve{t}$ and $\ve{r}$. We visualize some results at \figref{fig:results}(a-c).
Finally at bottom of \tabref{tbl:pose}, we list corresponding results on Dlake dataset, which draws similar conclusion with that from Zpark dataset.

% Finally, RNN gives strong cues about the moving speed and acceleration of the camera, yields the best results for both translation and rotation.

\medskip
\noindent\textbf{Segment Evaluation.}
At top part of \tabref{tbl:segment}, we show the scene parsing results of Zpark dataset.
Firstly, we adopt one of the SOTA parsing network on the CityScapes, \ie~ResNet38~\cite{wu2016wider}, and train it with Zpark dataset. It utilizes pre-trained parameters from the CityScapes~\cite{Cordts2016Cityscapes} dataset, and run with a 1.03$s$ per-frame with our resolution.
As shown at the 1st row, it achieve reasonable accuracy compare to our segment CNN (2nd row) when there is no pose priori. However, our network is 10x faster.
At 3rd row, we show the results of rendered label map with the estimated pose after pose RNN. Clearly, the results are much worse due to missing pixels and object misalignment.
At 4th row, we use the rendered label map with ground truth pose as segment CNN guidance to obtain an upper-bound for our segmentation performance.
In this case, the rendered label map aligns perfectly with the image, thus significantly improves the results by correct labelling most of the static background.
% At 4th row, we train the segment network without pose prior.
At 5th and 6th row, we show the results trained with rendered label map with pose after pose CNN and pose RNN respectively. We can see using pose CNN, the results just improve slightly compare to the segment CNN. From our observation, this is because the offset is still significant for some detailed structures, \eg light-pole.
However, when using the pose after RNN, better alignment is achieved, and the segment accuracy is improved significantly especially for thin structured regions like pole, as visualized in~\figref{fig:results}, which demonstrates the effectiveness of our strategy.

Bottom part of \tabref{tbl:segment} shows the results over the larger Dlake dataset with more object labelling, where we see clearer improvement, \ie from $62.36$ to $67.00$, and here the rendered label provides a background context for object segmentation, which also improve the object parsing performance.  In all classes, we observe the performance of traffic-light drops. In our opinion, the majorly reason is traffic-light only exists in intersection of roads, which happens much fewer than objects such as light-pole, yielding overfitting to the projected label maps from pose. We may fix this issue by training with even larger dataset or better class balancing strategies, which is left to our future work.

In \figref{fig:results}, we visualize several examples from our results at the view of camera. In the figure, we can see the noisy pose (a), is progressively rectified by pose CNN (b) and pose RNN (c) from view of camera.
Additionally, at (d) and (e), we compare the segment results without and with camera pose respectively. As can be seen at the boxed regions, the segment results with rendered label maps provide better accuracy in terms of capturing region details at the boundary, discovering rare classes and keeping correct scene layout. All of above could be important for applications, \eg~figuring out the traffic signs and tele-poles that are visually hard to detect. For additional visualization, please check our demo videos online~\footnote{Zpark: \url{https://www.youtube.com/watch?v=fqglYBipNfQ}}\footnote{Dlake \url{https://www.youtube.com/watch?v=fqglYBipNfQ}}.

% Given 3D point clouds, 2D pixel and instance-level annotations, background depth maps, camera pose information, a number of tasks could be defined.
% In current release, we mainly focus on the 2D image parsing task. We would like to add more tasks in near future.

\subsection{Benchmarks and baselines} \label{subsec:benchmark}
With various tasks and large amount of labelled data we have proposed, it would be non-practical for us to extensively explore algorithms over all of them. Therefore, we release the data to research community, and set up standard evaluation benchmarks.
Currently, four challenges have been set up online for evaluation by withholding part of our labelled results as test set, which include semantic segmentation~\cite{segment}, instance segmentation~\cite{instance}, self-localization~\cite{localization}, lanemark segmentation~\cite{lanemark}.

For evaluation, in the tasks of semantic segmentation, lanemark segmentation, we adopt mean IoU, and in the task of self-localization, we adopt median translation and rotation offset, which are described in evaluation of DeLS-3D (Sec.~\ref{subsec:dels-3d}).
For the task of instance segmentation, we use interpolated average precision (AP)~\cite{hariharan2014simultaneous} under various IoU thresholds which is used for the COCO challenge~\cite{lin2014microsoft}. % Due to it is possible that some instances in an image are labelled such as regions
Later, we elaborate the split of each dataset, the leading method on each benchmark currently.

\medskip
\noindent\textbf{Semantic segmentation.}
For video semantic segmentation, until now, we haven't receive valid results from the challenge. This probably is due to the extremely large amount of training videos in ApolloScape, making training a model with SOTA deep learning models such as ResNet~\cite{he2016deep} not-practical.  Thus, we select a subset from the whole data for comparison of one model performance between ApolloScape and Cityscapes. Specifically, 5,378 training images and 671 testing images are carefully selected from our 140K labelled semantic video frames for setting up the benchmark, which maintains the diversity and objects appeared of the collected scenes. The selected images will be released at our website~\cite{segment}.

We conducted our experiments using ResNet-38 network~\cite{wu2016wider} that trades depth for width comparing with the original ResNet structure~\cite{he2016deep}.
We fine-tune their released model using our training with initial learning rate 0.0001, standard SGD with momentum 0.9 and weight decay 0.0005, random crop with size $512\times512$, 10 times data augmentation that includes scaling and left-right flipping, and we train the network for 100 epochs.
The predictions are computed with the original image resolution without any post-processing steps such as multi-scale ensemble etc..
% To be comparable with the training and testing in the ResNet-38 network, we select a small subset from our dataset that consists of  which are at the same order of fine labeled images in the Cityscapes dataset (i.e., around 5K training images and 500 test images).
Tab.~\ref{tb:resnet38} shows the parsing results of classes in common for these two datasets. Notice that using exactly same training procedure, the test IoU with our dataset are much lower than that from the Cityscapes mostly due to the challenges we have mentioned at Sec.~\ref{subsec:spec}, especially for movable objects, where mIoU is 34.6\% lower than the corresponding one for the Cityscapes.

Here, we leave the training a model with the our full dataset to the research community and our future work.% when stronger computation resources are ready.

\begin{table}[t]\clearpage
\centering
\begin{tabular}{ l l c c }
  \toprule
  & & \multicolumn{2}{c}{IoU} \\
  Group & Class & Cityscapes & ApolloScape  \\
  \midrule
  movable & car & 94.67 & 87.12 \\
  object & motorcycle & 70.51 & 27.99 \\
  & bicycle & 78.55 & 48.65 \\
  & person & 83.17 & 57.12 \\
  & rider & 65.45 & 6.58 \\
  & truck & 62.43 & 25.28 \\
  & bus & 88.61 & 48.73  \\
  \midrule
  \textbf{mIoU} & & 77.63 & 43.07 \\
  \midrule
  surface & road & 97.94 & 92.03 \\
  & sidewalk & 84.08 & 46.42 \\
  \midrule
  infrastructure & fence & 61.49 & 42.08 \\
  & traffic light & 70.98 & 67.49 \\
  & pole & 62.11& 46.02 \\
  & traffic sign & 78.93 & 79.60 \\
  & wall & 58.81 & 8.41 \\
  & building & 92.66 & 65.71 \\
  \midrule
  nature & vegetation & 92.41 & 90.53 \\
  \bottomrule
\end{tabular}
\caption{Results of image parsing based on ResNet-38 network.}
\label{tb:resnet38}
\vspace{-1.0\baselineskip}
\end{table}

\begin{table*}[t]
\center
\fontsize{8}{7}\selectfont
\setlength\tabcolsep{1.5pt}
\setlength\extrarowheight{4pt}
\begin{tabular}{lc|ccccccccccccccccc}
Method & mIOU & \rotf{s\_w\_d} & \rotf{s\_y\_d} & \rotf{ds\_y\_dn} & \rotf{b\_w\_g} & \rotf{b\_y\_g} & \rotf{s\_w\_s} & \rotf{s\_w\_p} & \rotf{c\_wy\_z} & \rotf{a\_w\_t} & \rotf{a\_w\_tl} &\rotf{a\_w\_tr} & \rotf{a\_w\_l} & \rotf{a\_w\_r} & \rotf{a\_w\_lr} & \rotf{b\_n\_sr} & \rotf{r\_wy\_np} & \rotf{om\_n\_n} \\
\hline
ResNet38~\cite{wu2016wider} & 40.0 & 48.6 & 53.1 &57.8 & 52.1 & 22.7 &36.4 &18.7 &59.1 &40.4 &27.1 &49.1 & 57.4 & 20.9 & 0.01 & 0.9 & 36.1 & 40.5 \\
\toprule[0.1 em]
\end{tabular}
\caption{The IoU using one SOTA semantic segmentation architecture, \ie ResNet38~\cite{wu2016wider}. %Here,``s\_w\_d'' is short for solid, white and dividing in Tab.~\ref{tb:class2} by combining the first letter of type, color and usage respectively, and other classes are named accordingly.
Here the amount of class is less than that in Tab.~\ref{tb:class2} due to zero accuracy over some predefined labels.}
\label{tbl:lanemark}
\end{table*}

% \begin{table}[t]\clearpage
% \centering
% \begin{tabular}{ l l c c }
%   \toprule
% % \toprule
% % % metric & $1_{st}$~\cite{1st}  & $2_{nd}$~\cite{2nd} & $3_{nd}$~\cite{3rd} \\
% % \midrule
% mAP & 33.97 & 30.22 & 25.02
% \end{tabular}
% \caption{Top 3 results of instance segmentation based on the challenge leaderboard\footnotemark[\ref{link_inst}].}
% \label{tb:inst}
% \end{table}

\medskip
\noindent\textbf{Instance segmentation.} This task is an extension of semantic object parsing by jointly considering detection and segmentation. Specifically, we select 39212 training images and 1907 testing images, and set up a challenge benchmark online\cite{instance} evaluating 7 objects in our dataset (Upper part of Tab.~\ref{tb:resnet38}) to collect potential issues within autonomous driving scenario.
During the past few month, there are over 140 teams attended our challenge, which reveals our community is much more interested in object level understanding rather than scene segmentation.

The leading results from our participants are shown in Tab.~\ref{tb:inst}, where we can see in general the reported mAP of winning teams are lower than those reported in Cityscapes benchmarks~\cite{cityscapebenchmark}, by using similar strategies~\cite{liu2018path} modified from MaskRCNN~\cite{he2018mask}.
Based on the challenge reports from the winning team~\cite{1st}, comparing to Cityscapes, ApolloScape contains more tiny and occluded objects (60$\%$ object has scale less than $32$ pixels), which leads to significant drop of performance when transfer models trained on other datasets.

% \begin{table*}[t]
% \center
% \begin{tabular}{lc|cccccccccc}
% \toprule
% Method & mIOU & \rotf{s\_w\_d} & \rotf{s\_y\_d} & \rotf{ds\_y\_dn} & \rotf{b\_w\_g} & \rotf{b\_y\_g} & \rotf{s\_w\_s} & \rotf{s\_w\_p} & \rotf{c\_wy\_z} & \rotf{a\_w\_t} & \rotf{a\_w\_tl} \\
% \cmidrule{1-12}
% \multirow{4}{*}{ResNet38~\cite{wu2016wider}} & \multirow{4}{*}{40.0} & 48.6 & 53.1 &57.8 & 52.1 & 22.7 &36.4 &18.7 &59.1 &40.4 &27.1 \\
% \cmidrule{3-12}
% &      &\rotf{a\_w\_tr} & \rotf{a\_w\_l} & \rotf{a\_w\_r} & \rotf{a\_w\_lr} & \rotf{b\_n\_sr} & \rotf{r\_wy\_np} & \rotf{om\_n\_n}  & & \\
% \cmidrule{3-12}
% &      &49.1 & 57.4 & 20.9 & 0.01 & 0.9 & 36.1 & 40.5 & & \\
% \toprule[0.1 em]
% \end{tabular}
% \caption{The IoU over different classes using one SOTA semantic segmentation architecture, \ie ResNet38~\cite{wu2016wider}. Here,``s\_w\_d'' is short for solid, white and dividing in Tab.~\ref{tb:class2} by combining the first letter of type, color and usage respectively, and other classes are named accordingly. Here the amount of class is less than that in Tab.~\ref{tb:class2} due to zero accuracy over some predefined labels.}
% \label{tbl:lanemark}
% \end{table*}

\medskip
\noindent\textbf{Lanemark segmentation.} Lanemark segmentation task follows the same metric as semantic segmentation, which contains 132189 training images and 33790 testing images. Our in-house challenge benchmark \cite{lanemark} chooses to evaluate 35 most common lane mark types on the road as listed in Tab.~\ref{tb:class2}.

Until the submission of this paper, we only have one work based on ResNet-38 network~\cite{wu2016wider} evaluated, probably due to the large amount of data (160K+ images).
We show the corresponding detailed results in Tab.~\ref{tbl:lanemark}, where we can see the mIoU of each class are still very limited ($40\%$) comparing to the accuracy of leading semantic segmentation algorithms on general classes. We think this is mostly because the high contrast, dimmed and broken lane marks on the road such as the cases shown in Fig.~\ref{fig:scene}. % We wish in the near future, more research could be evolved in this task to improve the visual perception.

\begin{table}[t]
\centering
\begin{tabular}{l| l |l l l }
  \toprule
  Dataset & metric & $1_{st}$& $2_{nd}$& $3_{nd}$ \\
  \midrule
  ApolloScape & \parbox[t]{2mm}{\multirow{2}{*}{mAP}}  & 33.97~\cite{1st} & 30.22~\cite{2nd} & 25.02~\cite{3rd} \\
  \cmidrule{1-1}\cmidrule{3-5}
  Cityscapes &  & 38.0 & 37.2 & 36.4~\cite{liu2018path} \\
  \bottomrule
\end{tabular}
\caption{Results of top ranked instance segmentation algorithms in ApolloScape and Cityscapes (Numbers are obtained at the date of submission). }
\label{tb:inst}
\end{table}

\begin{table}[t]\clearpage
\centering
\begin{tabular}{ l c c }
  \toprule
%   & \multicolumn{2}{c}{}\\
%       \cmidrule{5-8}
  Road ID &  Trans (m) $\downarrow$ & Rot ($\circ$)$\downarrow$ \\
  \midrule
  Road11 & 0.0476 & 0.0452 \\
  Road12 & 0.1115 & 0.0528 \\
  Road14 & 0.0785 & 0.0825 \\
  Road15 & 0.0711 & 0.1240 \\
  Road16 & 0.1229 & 0.2063 \\
  Road17 & 0.4934 & 0.3135 \\
  \midrule
  mean & 0.1542 & 0.1374 \\
  \bottomrule
\end{tabular}
\caption{Detailed localization accuracy of the leading results on our benchmark from~\cite{liu2018deep}.}
\label{tbl:self-loc}
\vspace{-1.1\baselineskip}
\end{table}

\medskip
\noindent\textbf{Self-localization.} We use the same metrics for evaluating camera pose, \ie median offset of translation and rotation, as described in Sec.~\ref{subsec:dels-3d}. This task contains driving videos in 6 sites from Beijing, Guangzhou, Chengdu and Shanghai in China, under multiple driving scenarios and day times. In total, we provide 153 training videos and 71 testing videos including over 300k image frames, and build an in-house challenge benchmark website \cite{localization} most recently.

Currently, we also have few submissions, while the leading one published is from one of the SOTA method for large-scale image based localization~\cite{liu2018deep}. The method is based on image retrieval with learned deep features via various triplet losses. We show their reported number in Tab.~\ref{tbl:self-loc}, where the localization errors are surprisingly small, \ie~translation is around 15cm and rotation error is around 0.14 degree. Originally, we believe image appearance similarity on the street or highway can fail deep network models. However, from the participant results, especially designed features distinguish minor appearance changes and provide high accurate localization results. Another possibility is that our acquisition vehicle always drives in a roughly constant speed, reducing the issues from speed changing in real applications. In the near future, hopefully, we can add more challenging scenarios with more variations in driving speed and weathers.

In summary, from the dataset benchmarks we set up and evaluated algorithms, we found for low-level localization, the results are impressively good, while for high level semantic understanding, Apolloscape provides additional challenges and new issues, yielding limited accuracy for SOTA algorithms, \ie~best mAP is around $33\%$ for instance segmentation, and best mIoU is around $40\%$ for lane segmentation. Comparing to human perception, visual based algorithms for autonomous driving definitely need further research to handle extremely difficult cases.

\section{Conclusion and Future Work}
In this paper, we present the \emph{ApolloScape}, a large, diverse, and multi-task dataset for autonomous driving, which includes high density 3D point cloud map, per-pixel, per-frame semantic image label, lane mark label, semantic instance segmentation for various videos. Every frame of our videos is geo-tagged with high accurate GPS/IMU device. \emph{ApolloScape} is significantly larger than existing autonomous driving datasets, \eg~ KITTI~\cite{Geiger2013IJRR} and Cityscapes~\cite{Cordts2016Cityscapes}, yielding more challenges for computer vision research field.
In order to label such a large dataset, we developed an active 2D/3D joint annotation pipeline, which effectively accelerate the labelling process.  Back on \emph{ApolloScape}, we developed a joint localization and segmentation algorithm with a 3D semantic map, which fuses multi-sensors, is simple and runs efficiently, yielding strong results in both tasks. We hope it may motivate researcher to develop algorithms handling multiple tasks simultaneously by considering their inner geometrical relationships. Finally, for each individual task, we set up an online evaluation benchmark where different algorithms can compete with a fair platform.

Last but not the least, \emph{ApolloScape} is an evolving dataset, not only in terms of data scale, but also in terms of various driving conditions, tasks and acquisition devices. For example, firstly, we plan to enlarge our dataset to contain more diversified driving environments including snow, and foggy. Secondly, we \peng{also released our labelled 3D cars~\cite{songApolloCar3D}, stereo images, 3D humans and tracking~\cite{ma2018trafficpredict} of objects in 3D recently}.  Thirdly, we plan to mount a panoramic camera system, and Velodyne~\cite{velodyne} in near future to generate depth maps for objects and panoramic images. % In the current release, the depth information for the moving objects is still missing. We would like to produce complete depth information for both static background and moving objects.
\ifCLASSOPTIONcompsoc
  \section*{Acknowledgments}
\else
  \section*{Acknowledgment}
\fi
This work is supported by Baidu Inc.. We also thank the work of Xibin Song, Binbin Cao, Jin Fang, He Jiang, Yu Zhang, Xiang Gu, and Xiaofei Liu for their laborious efforts in organizing data,  helping writing label tools, checking labelled results and manage the content of benchmark websites. We thank Alan L. Yuille, Hongdong Li and Andreas Geiger for benchmark suggestions.

\ifCLASSOPTIONcaptionsoff
  \newpage
\fi



\bibliographystyle{IEEEtran}
\bibliography{egbib}

\begin{thebibliography}{10}
\providecommand{\url}[1]{#1}
\csname url@samestyle\endcsname
\providecommand{\newblock}{\relax}
\providecommand{\bibinfo}[2]{#2}
\providecommand{\BIBentrySTDinterwordspacing}{\spaceskip=0pt\relax}
\providecommand{\BIBentryALTinterwordstretchfactor}{4}
\providecommand{\BIBentryALTinterwordspacing}{\spaceskip=\fontdimen2\font plus
\BIBentryALTinterwordstretchfactor\fontdimen3\font minus
  \fontdimen4\font\relax}
\providecommand{\BIBforeignlanguage}[2]{{%
\expandafter\ifx\csname l@#1\endcsname\relax
\typeout{** WARNING: IEEEtran.bst: No hyphenation pattern has been}%
\typeout{** loaded for the language `#1'. Using the pattern for}%
\typeout{** the default language instead.}%
\else
\language=\csname l@#1\endcsname
\fi
#2}}
\providecommand{\BIBdecl}{\relax}
\BIBdecl

\bibitem{apolloweb}
``{ApolloScape Website},'' \url{apolloscape.auto}.

\bibitem{Geiger2013IJRR}
A.~Geiger, P.~Lenz, C.~Stiller, and R.~Urtasun, ``Vision meets robotics: The
  kitti dataset,'' \emph{International Journal of Robotics Research (IJRR)},
  2013.

\bibitem{Cordts2016Cityscapes}
M.~Cordts, M.~Omran, S.~Ramos, T.~Rehfeld, M.~Enzweiler, R.~Benenson,
  U.~Franke, S.~Roth, and B.~Schiele, ``The cityscapes dataset for semantic
  urban scene understanding,'' in \emph{Proc. of the IEEE Conference on
  Computer Vision and Pattern Recognition (CVPR)}, 2016.

\bibitem{velodyne}
{Velodyne Lidar}, ``{HDL-64E},'' \url{http://velodynelidar.com/}, 2018,
  [Online; accessed 01-March-2018].

\bibitem{kar2017learning}
A.~Kar, C.~H{\"a}ne, and J.~Malik, ``Learning a multi-view stereo machine,'' in
  \emph{Advances in neural information processing systems}, 2017, pp. 365--376.

\bibitem{huang2018deepmvs}
P.-H. Huang, K.~Matzen, J.~Kopf, N.~Ahuja, and J.-B. Huang, ``Deepmvs: Learning
  multi-view stereopsis,'' in \emph{Proceedings of the IEEE Conference on
  Computer Vision and Pattern Recognition}, 2018, pp. 2821--2830.

\bibitem{Yao_2018_ECCV}
Y.~Yao, Z.~Luo, S.~Li, T.~Fang, and L.~Quan, ``Mvsnet: Depth inference for
  unstructured multi-view stereo,'' in \emph{The European Conference on
  Computer Vision (ECCV)}, September 2018.

\bibitem{cheng2018depth}
X.~Cheng, P.~Wang, and R.~Yang, ``Depth estimation via affinity learned with
  convolutional spatial propagation network,'' \emph{European Conference on
  Computer Vision}, 2018.

\bibitem{Kendall_2015_ICCV}
A.~Kendall, M.~Grimes, and R.~Cipolla, ``Posenet: A convolutional network for
  real-time 6-dof camera relocalization,'' in \emph{Proceedings of the IEEE
  international conference on computer vision}, 2015, pp. 2938--2946.

\bibitem{schonberger2018semantic}
J.~L. Sch{\"o}nberger, M.~Pollefeys, A.~Geiger, and T.~Sattler, ``Semantic
  visual localization,'' \emph{ISPRS Journal of Photogrammetry and Remote
  Sensing (JPRS)}, 2018.

\bibitem{long2015fully}
J.~Long, E.~Shelhamer, and T.~Darrell, ``Fully convolutional networks for
  semantic segmentation,'' in \emph{Proceedings of the IEEE conference on
  computer vision and pattern recognition}, 2015, pp. 3431--3440.

\bibitem{ChenPSA17}
L.~Chen, G.~Papandreou, F.~Schroff, and H.~Adam, ``Rethinking atrous
  convolution for semantic image segmentation,'' \emph{CoRR}, vol.
  abs/1706.05587, 2017.

\bibitem{he2018mask}
K.~He, G.~Gkioxari, P.~Doll{\'a}r, and R.~Girshick, ``Mask r-cnn,'' \emph{IEEE
  transactions on pattern analysis and machine intelligence}, 2018.

\bibitem{chen2017masklab}
L.-C. Chen, A.~Hermans, G.~Papandreou, F.~Schroff, P.~Wang, and H.~Adam,
  ``Masklab: Instance segmentation by refining object detection with semantic
  and direction features,'' \emph{arXiv preprint arXiv:1712.04837}, 2017.

\bibitem{xiang2014beyond}
Y.~Xiang, R.~Mottaghi, and S.~Savarese, ``Beyond pascal: A benchmark for 3d
  object detection in the wild,'' in \emph{Applications of Computer Vision
  (WACV), 2014 IEEE Winter Conference on}.\hskip 1em plus 0.5em minus
  0.4em\relax IEEE, 2014, pp. 75--82.

\bibitem{kar2015category}
A.~Kar, S.~Tulsiani, J.~Carreira, and J.~Malik, ``Category-specific object
  reconstruction from a single image,'' in \emph{Proceedings of the IEEE
  Conference on Computer Vision and Pattern Recognition}, 2015, pp. 1966--1974.

\bibitem{guney2015displets}
F.~Guney and A.~Geiger, ``Displets: Resolving stereo ambiguities using object
  knowledge,'' in \emph{Proceedings of the IEEE Conference on Computer Vision
  and Pattern Recognition}, 2015, pp. 4165--4175.

\bibitem{kundu20183d}
A.~Kundu, Y.~Li, and J.~M. Rehg, ``3d-rcnn: Instance-level 3d object
  reconstruction via render-and-compare,'' in \emph{Proceedings of the IEEE
  Conference on Computer Vision and Pattern Recognition}, 2018, pp. 3559--3568.

\bibitem{songApolloCar3D}
X.~Song, P.~Wang, D.~Zhou, R.~Zhu, C.~Guan, Y.~Dai, H.~Su, H.~Li, and R.~Yang,
  ``Apollocar3d: {A} large 3d car instance understanding benchmark for
  autonomous driving,'' \emph{CVPR}, 2019.

\bibitem{segment}
ApolloScape., ``Semantic segmentation,''
  \url{http://apolloscape.auto/scene.html}.

\bibitem{instance}
------, ``Instance segmentation,''
  \url{https://www.kaggle.com/c/cvpr-2018-autonomous-driving}.

\bibitem{lanemark}
------, ``Lanemark segmentation,''
  \url{http://apolloscape.auto/lane_segmentation.html}.

\bibitem{localization}
------, ``Localization,'' \url{http://apolloscape.auto/self_localization.html}.

\bibitem{apolloapi}
W.~Peng \emph{et~al.}, ``{ApolloScape API},''
  \url{https://github.com/ApolloScapeAuto/dataset-api}.

\bibitem{kundu2014joint}
A.~Kundu, Y.~Li, F.~Dellaert, F.~Li, and J.~M. Rehg, ``Joint semantic
  segmentation and 3d reconstruction from monocular video,'' in \emph{European
  Conference on Computer Vision}.\hskip 1em plus 0.5em minus 0.4em\relax
  Springer, 2014, pp. 703--718.

\bibitem{brostow2009semantic}
G.~J. Brostow, J.~Fauqueur, and R.~Cipolla, ``Semantic object classes in video:
  A high-definition ground truth database,'' \emph{Pattern Recognition
  Letters}, vol.~30, no.~2, pp. 88--97, 2009.

\bibitem{wang2017torontocity}
S.~Wang, M.~Bai, G.~Mattyus, H.~Chu, W.~Luo, B.~Yang, J.~Liang, J.~Cheverie,
  S.~Fidler, and R.~Urtasun, ``Torontocity: Seeing the world with a million
  eyes,'' in \emph{Proceedings of the IEEE Conference on Computer Vision and
  Pattern Recognition}, 2017, pp. 3009--3017.

\bibitem{neuhold2017mapillary}
G.~Neuhold, T.~Ollmann, S.~R. Bulo, and P.~Kontschieder, ``The mapillary vistas
  dataset for semantic understanding of street scenes,'' in \emph{Proceedings
  of the International Conference on Computer Vision (ICCV), Venice, Italy},
  2017, pp. 22--29.

\bibitem{yu2018bdd100k}
F.~Yu, W.~Xian, Y.~Chen, F.~Liu, M.~Liao, V.~Madhavan, and T.~Darrell,
  ``Bdd100k: A diverse driving video database with scalable annotation
  tooling,'' \emph{arXiv preprint arXiv:1805.04687}, 2018.

\bibitem{RosCVPR16}
G.~Ros, L.~Sellart, J.~Materzynska, D.~Vazquez, and A.~M. Lopez, ``The synthia
  dataset: A large collection of synthetic images for semantic segmentation of
  urban scenes,'' in \emph{Proceedings of the IEEE conference on computer
  vision and pattern recognition}, 2016, pp. 3234--3243.

\bibitem{richter2017playing}
S.~R. Richter, Z.~Hayder, and V.~Koltun, ``Playing for benchmarks,'' in
  \emph{International Conference on Computer Vision (ICCV)}, 2017.

\bibitem{scharstein2014high}
D.~Scharstein, H.~Hirschm{\"u}ller, Y.~Kitajima, G.~Krathwohl,
  N.~Ne{\v{s}}i{\'c}, X.~Wang, and P.~Westling, ``High-resolution stereo
  datasets with subpixel-accurate ground truth,'' in \emph{German Conference on
  Pattern Recognition}.\hskip 1em plus 0.5em minus 0.4em\relax Springer, 2014,
  pp. 31--42.

\bibitem{silberman2012indoor}
N.~Silberman, D.~Hoiem, P.~Kohli, and R.~Fergus, ``Indoor segmentation and
  support inference from rgbd images,'' in \emph{European Conference on
  Computer Vision}.\hskip 1em plus 0.5em minus 0.4em\relax Springer, 2012, pp.
  746--760.

\bibitem{sattler2018benchmarking}
T.~Sattler, W.~Maddern, C.~Toft, A.~Torii, L.~Hammarstrand, E.~Stenborg,
  D.~Safari, M.~Okutomi, M.~Pollefeys, J.~Sivic \emph{et~al.}, ``Benchmarking
  6dof outdoor visual localization in changing conditions,'' in
  \emph{Proceedings of the IEEE Conference on Computer Vision and Pattern
  Recognition}, vol.~1, 2018.

\bibitem{everingham2010pascal}
M.~Everingham, L.~Van~Gool, C.~K. Williams, J.~Winn, and A.~Zisserman, ``The
  pascal visual object classes (voc) challenge,'' \emph{International journal
  of computer vision}, vol.~88, no.~2, pp. 303--338, 2010.

\bibitem{lin2014microsoft}
T.-Y. Lin, M.~Maire, S.~Belongie, J.~Hays, P.~Perona, D.~Ramanan,
  P.~Doll{\'a}r, and C.~L. Zitnick, ``Microsoft coco: Common objects in
  context,'' in \emph{European conference on computer vision}.\hskip 1em plus
  0.5em minus 0.4em\relax Springer, 2014, pp. 740--755.

\bibitem{unity}
``{Unity Development Platform},'' \url{https://unity3d.com/}.

\bibitem{hoffman2016fcns}
J.~Hoffman, D.~Wang, F.~Yu, and T.~Darrell, ``Fcns in the wild: Pixel-level
  adversarial and constraint-based adaptation,'' \emph{arXiv preprint
  arXiv:1612.02649}, 2016.

\bibitem{zhang2017curriculum}
Y.~Zhang, P.~David, and B.~Gong, ``Curriculum domain adaptation for semantic
  segmentation of urban scenes,'' in \emph{The IEEE International Conference on
  Computer Vision (ICCV)}, vol.~2, no.~5, 2017, p.~6.

\bibitem{chen2018road}
Y.~Chen, W.~Li, and L.~Van~Gool, ``Road: Reality oriented adaptation for
  semantic segmentation of urban scenes,'' in \emph{Proceedings of the IEEE
  Conference on Computer Vision and Pattern Recognition}, 2018, pp. 7892--7901.

\bibitem{haralick1994review}
B.~M. Haralick, C.-N. Lee, K.~Ottenberg, and M.~N{\"o}lle, ``Review and
  analysis of solutions of the three point perspective pose estimation
  problem,'' \emph{IJCV}, vol.~13, no.~3, pp. 331--356, 1994.

\bibitem{kneip2014upnp}
L.~Kneip, H.~Li, and Y.~Seo, ``Upnp: An optimal o (n) solution to the absolute
  pose problem with universal applicability,'' in \emph{European Conference on
  Computer Vision}.\hskip 1em plus 0.5em minus 0.4em\relax Springer, 2014, pp.
  127--142.

\bibitem{david2004softposit}
P.~David, D.~Dementhon, R.~Duraiswami, and H.~Samet, ``Softposit: Simultaneous
  pose and correspondence determination,'' \emph{IJCV}, vol.~59, no.~3, pp.
  259--284, 2004.

\bibitem{moreno2008pose}
F.~Moreno-Noguer, V.~Lepetit, and P.~Fua, ``Pose priors for simultaneously
  solving alignment and correspondence,'' \emph{European Conference on Computer
  Vision}, pp. 405--418, 2008.

\bibitem{campbell2017globally}
D.~Campbell, L.~Petersson, L.~Kneip, and H.~Li, ``Globally-optimal inlier set
  maximisation for simultaneous camera pose and feature correspondence,'' in
  \emph{The IEEE International Conference on Computer Vision (ICCV)}, vol.~1,
  no.~3, 2017.

\bibitem{sattler2017large}
T.~Sattler, A.~Torii, J.~Sivic, M.~Pollefeys, H.~Taira, M.~Okutomi, and
  T.~Pajdla, ``Are large-scale 3d models really necessary for accurate visual
  localization?'' in \emph{2017 IEEE Conference on Computer Vision and Pattern
  Recognition (CVPR)}.\hskip 1em plus 0.5em minus 0.4em\relax IEEE, 2017, pp.
  6175--6184.

\bibitem{engel2014lsd}
J.~Engel, T.~Sch{\"o}ps, and D.~Cremers, ``Lsd-slam: Large-scale direct
  monocular slam,'' in \emph{European Conference on Computer Vision}.\hskip 1em
  plus 0.5em minus 0.4em\relax Springer, 2014, pp. 834--849.

\bibitem{kendall2017geometric}
A.~Kendall, R.~Cipolla \emph{et~al.}, ``Geometric loss functions for camera
  pose regression with deep learning,'' in \emph{Proceedings of the IEEE
  Conference on Computer Vision and Pattern Recognition (CVPR)}, vol.~3, 2017,
  p.~8.

\bibitem{hazirbasimage}
F.~Walch, C.~Hazirbas, L.~Leal-Taixe, T.~Sattler, S.~Hilsenbeck, and
  D.~Cremers, ``Image-based localization using lstms for structured feature
  correlation,'' in \emph{Int. Conf. Comput. Vis.(ICCV)}, 2017, pp. 627--637.

\bibitem{DBLP:journals/corr/ClarkWMTW17}
R.~Clark, S.~Wang, A.~Markham, N.~Trigoni, and H.~Wen, ``Vidloc: A deep
  spatio-temporal model for 6-dof video-clip relocalization,'' in
  \emph{Proceedings of the IEEE Conference on Computer Vision and Pattern
  Recognition (CVPR)}, vol.~3, 2017.

\bibitem{coskun2017long}
H.~Coskun, F.~Achilles, R.~DiPietro, N.~Navab, and F.~Tombari, ``Long
  short-term memory kalman filters: Recurrent neural estimators for pose
  regularization,'' in \emph{Proceedings of the International Conference on
  Computer Vision (ICCV)}, 2017.

\bibitem{lianos2018vso}
K.-N. Lianos, J.~L. Sch{\"o}nberger, M.~Pollefeys, and T.~Sattler, ``Vso:
  Visual semantic odometry,'' in \emph{Proceedings of the European Conference
  on Computer Vision (ECCV)}, 2018, pp. 234--250.

\bibitem{vishal2015accurate}
K.~Vishal, C.~Jawahar, and V.~Chari, ``Accurate localization by fusing images
  and gps signals,'' in \emph{Proceedings of the IEEE Conference on Computer
  Vision and Pattern Recognition Workshops}, 2015, pp. 17--24.

\bibitem{laskar2017camera}
Z.~Laskar, I.~Melekhov, S.~Kalia, and J.~Kannala, ``Camera relocalization by
  computing pairwise relative poses using convolutional neural network,''
  \emph{Proceedings of the IEEE Conference on Computer Vision and Pattern
  Recognition}, 2017.

\bibitem{ummenhofer2016demon}
B.~Ummenhofer, H.~Zhou, J.~Uhrig, N.~Mayer, E.~Ilg, A.~Dosovitskiy, and
  T.~Brox, ``Demon: Depth and motion network for learning monocular stereo,''
  in \emph{IEEE Conference on computer vision and pattern recognition (CVPR)},
  vol.~5, 2017, p.~6.

\bibitem{ZhaoSQWJ16}
H.~Zhao, J.~Shi, X.~Qi, X.~Wang, and J.~Jia, ``Pyramid scene parsing network,''
  \emph{CVPR}, 2017.

\bibitem{higherordercrf_ECCV2016}
A.~Arnab, S.~Jayasumana, S.~Zheng, and P.~H. Torr, ``Higher order conditional
  random fields in deep neural networks,'' in \emph{European Conference on
  Computer Vision}.\hskip 1em plus 0.5em minus 0.4em\relax Springer, 2016, pp.
  524--540.

\bibitem{byeon2015scene}
W.~Byeon, T.~M. Breuel, F.~Raue, and M.~Liwicki, ``Scene labeling with lstm
  recurrent neural networks,'' in \emph{Proceedings of the IEEE Conference on
  Computer Vision and Pattern Recognition}, 2015, pp. 3547--3555.

\bibitem{he2016deep}
K.~He, X.~Zhang, S.~Ren, and J.~Sun, ``Deep residual learning for image
  recognition,'' in \emph{Proceedings of the IEEE conference on computer vision
  and pattern recognition}, 2016, pp. 770--778.

\bibitem{PaszkeCKC16}
A.~Paszke, A.~Chaurasia, S.~Kim, and E.~Culurciello, ``Enet: {A} deep neural
  network architecture for real-time semantic segmentation,'' \emph{CoRR}, vol.
  abs/1606.02147, 2016.

\bibitem{ZhaoQSSJ17}
H.~Zhao, X.~Qi, X.~Shen, J.~Shi, and J.~Jia, ``Icnet for real-time semantic
  segmentation on high-resolution images,'' \emph{CoRR}, vol. abs/1704.08545,
  2017.

\bibitem{kundu2016feature}
A.~Kundu, V.~Vineet, and V.~Koltun, ``Feature space optimization for semantic
  video segmentation,'' in \emph{Proceedings of the IEEE Conference on Computer
  Vision and Pattern Recognition}, 2016, pp. 3168--3175.

\bibitem{dosovitskiy2015flownet}
A.~Dosovitskiy, P.~Fischer, E.~Ilg, P.~Hausser, C.~Hazirbas, V.~Golkov, P.~Van
  Der~Smagt, D.~Cremers, and T.~Brox, ``Flownet: Learning optical flow with
  convolutional networks,'' in \emph{Proceedings of the IEEE International
  Conference on Computer Vision}, 2015, pp. 2758--2766.

\bibitem{gadde2017semantic}
R.~Gadde, V.~Jampani, and P.~V. Gehler, ``Semantic video cnns through
  representation warping,'' \emph{Proceedings of the International Conference
  on Computer Vision (ICCV)}, 2017.

\bibitem{zhu2016deep}
X.~Zhu, Y.~Xiong, J.~Dai, L.~Yuan, and Y.~Wei, ``Deep feature flow for video
  recognition,'' \emph{Proceedings of the IEEE Conference on Computer Vision
  and Pattern Recognition}, 2017.

\bibitem{hane2013joint}
C.~Hane, C.~Zach, A.~Cohen, R.~Angst, and M.~Pollefeys, ``Joint 3d scene
  reconstruction and class segmentation,'' in \emph{Proceedings of the IEEE
  Conference on Computer Vision and Pattern Recognition}, 2013, pp. 97--104.

\bibitem{tateno2017cnn}
K.~Tateno, F.~Tombari, I.~Laina, and N.~Navab, ``Cnn-slam: Real-time dense
  monocular slam with learned depth prediction,'' in \emph{Proceedings of the
  IEEE Conference on Computer Vision and Pattern Recognition (CVPR)}, vol.~2,
  2017.

\bibitem{rieglvmx}
RIEGL, ``{VMX-1HA},'' \url{http://www.riegl.com/}.

\bibitem{lanemark_tu}
TuSimple., ``Lanemark segmentation,'' \url{http://benchmark.tusimple.ai/\#/}.

\bibitem{revaud2016deepmatching}
J.~Revaud, P.~Weinzaepfel, Z.~Harchaoui, and C.~Schmid, ``Deepmatching:
  Hierarchical deformable dense matching,'' \emph{IJCV}, vol. 120, no.~3, pp.
  300--323, 2016.

\bibitem{luo2018every}
C.~Luo, Z.~Yang, P.~Wang, Y.~Wang, W.~Xu, R.~Nevatia, and A.~Yuille, ``Every
  pixel counts++: Joint learning of geometry and motion with 3d holistic
  understanding,'' \emph{arXiv preprint arXiv:1810.06125}, 2018.

\bibitem{xie2016semantic}
J.~Xie, M.~Kiefel, M.-T. Sun, and A.~Geiger, ``Semantic instance annotation of
  street scenes by 3d to 2d label transfer,'' in \emph{Proceedings of the IEEE
  Conference on Computer Vision and Pattern Recognition}, 2016, pp. 3688--3697.

\bibitem{christoph2014object}
S.~Christoph~Stein, M.~Schoeler, J.~Papon, and F.~Worgotter, ``Object
  partitioning using local convexity,'' in \emph{Proceedings of the IEEE
  Conference on Computer Vision and Pattern Recognition}, 2014, pp. 304--311.

\bibitem{pcl}
``{Point Cloud Library},'' \url{pointclouds.org}.

\bibitem{qi2017pointnet++}
C.~R. Qi, L.~Yi, H.~Su, and L.~J. Guibas, ``Pointnet++: Deep hierarchical
  feature learning on point sets in a metric space,'' in \emph{Advances in
  Neural Information Processing Systems}, 2017, pp. 5105--5114.

\bibitem{wu2016wider}
Z.~Wu, C.~Shen, and A.~v.~d. Hengel, ``Wider or deeper: Revisiting the resnet
  model for visual recognition,'' \emph{arXiv preprint arXiv:1611.10080}, 2016.

\bibitem{ren2015faster}
S.~Ren, K.~He, R.~Girshick, and J.~Sun, ``Faster r-cnn: Towards real-time
  object detection with region proposal networks,'' in \emph{Advances in neural
  information processing systems}, 2015, pp. 91--99.

\bibitem{wang2015joint}
P.~Wang, X.~Shen, Z.~Lin, S.~Cohen, B.~Price, and A.~L. Yuille, ``Joint object
  and part segmentation using deep learned potentials,'' in \emph{Proceedings
  of the IEEE International Conference on Computer Vision}, 2015, pp.
  1573--1581.

\bibitem{su2012crowdsourcing}
H.~Su, J.~Deng, and L.~Fei-Fei, ``Crowdsourcing annotations for visual object
  detection,'' in \emph{Workshops at the Twenty-Sixth AAAI Conference on
  Artificial Intelligence}, vol.~1, no.~2, 2012.

\bibitem{kovashka2016crowdsourcing}
A.~Kovashka, O.~Russakovsky, L.~Fei-Fei, K.~Grauman \emph{et~al.},
  ``Crowdsourcing in computer vision,'' \emph{Foundations and
  Trends{\textregistered} in Computer Graphics and Vision}, vol.~10, no.~3, pp.
  177--243, 2016.

\bibitem{li2016crowdsourced}
G.~Li, J.~Wang, Y.~Zheng, and M.~J. Franklin, ``Crowdsourced data management: A
  survey,'' \emph{IEEE Transactions on Knowledge and Data Engineering},
  vol.~28, no.~9, pp. 2296--2319, 2016.

\bibitem{wang2018dels}
P.~Wang, R.~Yang, B.~Cao, W.~Xu, and Y.~Lin, ``Dels-3d: Deep localization and
  segmentation with a 3d semantic map,'' in \emph{Proceedings of the IEEE
  Conference on Computer Vision and Pattern Recognition}, 2018, pp. 5860--5869.

\bibitem{wu2016google}
Y.~Wu, M.~Schuster, Z.~Chen, Q.~V. Le, M.~Norouzi, W.~Macherey, M.~Krikun,
  Y.~Cao, Q.~Gao, K.~Macherey \emph{et~al.}, ``Google's neural machine
  translation system: Bridging the gap between human and machine translation,''
  \emph{arXiv preprint arXiv:1609.08144}, 2016.

\bibitem{kalman1960new}
R.~E. Kalman \emph{et~al.}, ``A new approach to linear filtering and prediction
  problems,'' \emph{Journal of basic Engineering}, vol.~82, no.~1, pp. 35--45,
  1960.

\bibitem{pohlen2016full}
T.~Pohlen, A.~Hermans, M.~Mathias, and B.~Leibe, ``Full-resolution residual
  networks for semantic segmentation in street scenes,'' \emph{Proceedings of
  the IEEE Conference on Computer Vision and Pattern Recognition}, 2017.

\bibitem{eigen2015predicting}
D.~Eigen and R.~Fergus, ``Predicting depth, surface normals and semantic labels
  with a common multi-scale convolutional architecture,'' in \emph{Proceedings
  of the IEEE International Conference on Computer Vision}, 2015, pp.
  2650--2658.

\bibitem{lee2015gps}
B.-H. Lee, J.-H. Song, J.-H. Im, S.-H. Im, M.-B. Heo, and G.-I. Jee, ``Gps/dr
  error estimation for autonomous vehicle localization,'' \emph{Sensors},
  vol.~15, no.~8, pp. 20\,779--20\,798, 2015.

\bibitem{dozat2016incorporating}
T.~Dozat, ``Incorporating nesterov momentum into adam,'' 2016.

\bibitem{ChenLLLWWXXZZ15}
T.~Chen, M.~Li, Y.~Li, M.~Lin, N.~Wang, M.~Wang, T.~Xiao, B.~Xu, C.~Zhang, and
  Z.~Zhang, ``Mxnet: {A} flexible and efficient machine learning library for
  heterogeneous distributed systems,'' \emph{CoRR}, vol. abs/1512.01274, 2015.

\bibitem{murTRO2015}
R.~Mur-Artal, J.~M.~M. Montiel, and J.~D. Tardos, ``Orb-slam: a versatile and
  accurate monocular slam system,'' \emph{IEEE transactions on robotics},
  vol.~31, no.~5, pp. 1147--1163, 2015.

\bibitem{hariharan2014simultaneous}
B.~Hariharan, P.~Arbel{\'a}ez, R.~Girshick, and J.~Malik, ``Simultaneous
  detection and segmentation,'' in \emph{European Conference on Computer
  Vision}.\hskip 1em plus 0.5em minus 0.4em\relax Springer, 2014, pp. 297--312.

\bibitem{cityscapebenchmark}
``Cityscapes instance segmentation benchmark,''
  \url{https://www.cityscapes-dataset.com/benchmarks/\#instance-level-scene-labeling-task}.

\bibitem{liu2018path}
S.~Liu, L.~Qi, H.~Qin, J.~Shi, and J.~Jia, ``Path aggregation network for
  instance segmentation,'' in \emph{Proceedings of the IEEE Conference on
  Computer Vision and Pattern Recognition}, 2018, pp. 8759--8768.

\bibitem{1st}
Z.~Yueqing, L.~Zeming, and G.~Yu, ``Find tiny instance segmentation,''
  \url{http://www.skicyyu.org/WAD/wad_final.pdf}, 2018.

\bibitem{2nd}
Smart\_Vision\_SG, ``Wad instance segmentation 2nd place,''
  \url{https://github.com/Computational-Camera/Kaggle-CVPR-2018-WAD-Video-Segmentation-Challenge-Solution},
  2018.

\bibitem{3rd}
SZU\_N606, ``Wad instance segmentation 3rd place,''
  \url{https://github.com/wwoody827/cvpr-2018-autonomous-driving-autopilot-solution},
  2018.

\bibitem{liu2018deep}
L.~Liu, H.~Li, and Y.~Dai, ``Deep stochastic attraction and repulsion embedding
  for image based localization,'' \emph{arXiv preprint arXiv:1808.08779}, 2018.

\bibitem{ma2018trafficpredict}
Y.~Ma, X.~Zhu, S.~Zhang, R.~Yang, W.~Wang, and D.~Manocha, ``Trafficpredict:
  Trajectory prediction for heterogeneous traffic-agents,'' \emph{AAAI}, 2019.

\end{thebibliography}
%IEEEabrv


% \end{thebibliography}


\begin{IEEEbiography}[{\includegraphics[width=1in,height=1.25in,clip,keepaspectratio]{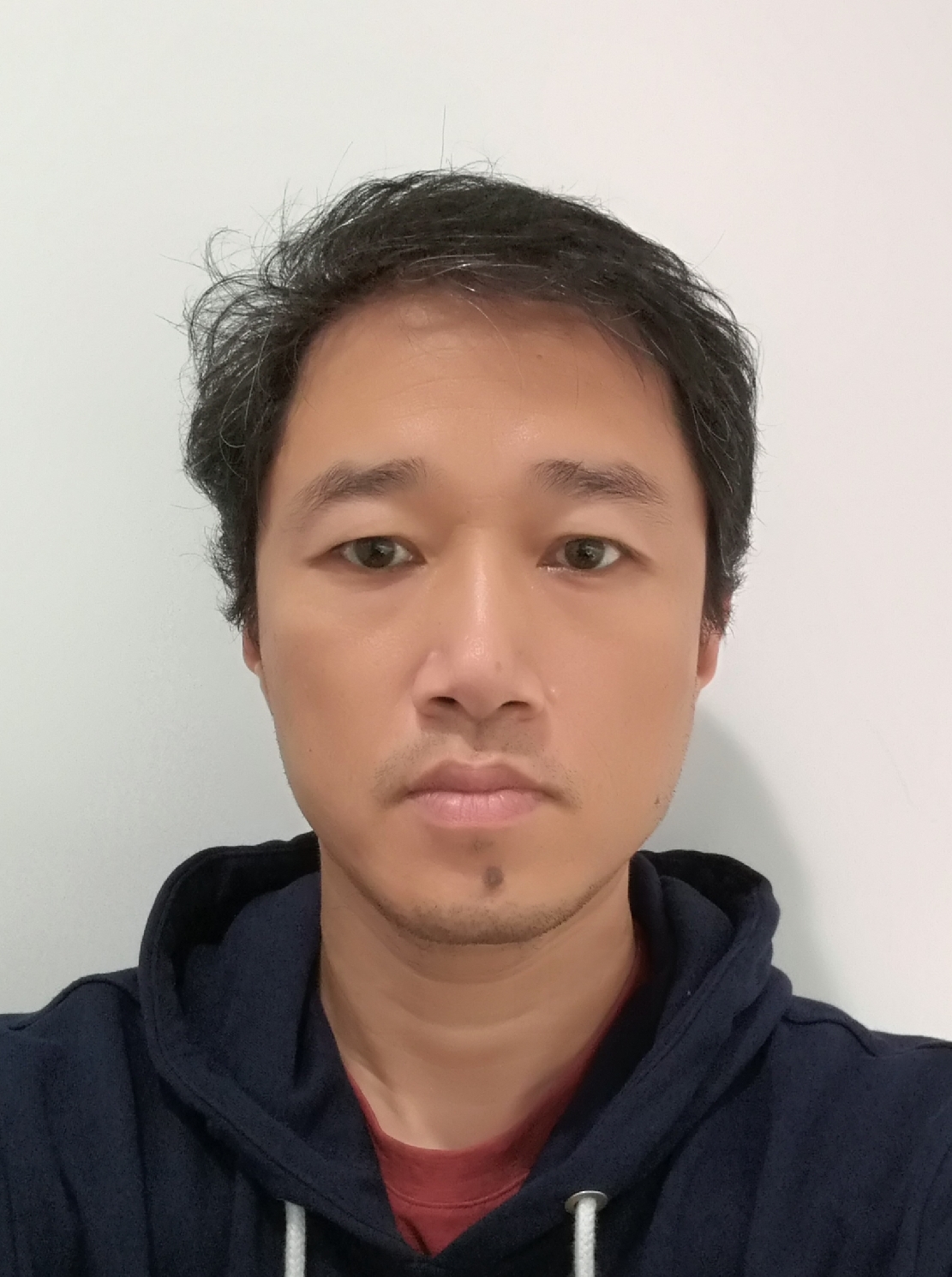}}]{Xinyu Huang}
is currently a senior research scientist at Baidu Research and an associate professor at the North Carolina Central University. He obtained his Ph.D. degree from the University of Kentucky and B.S. degree from Huazhong University of Science and Technology. His research areas include computer vision, image processing, pattern recognition, and machine learning.
\end{IEEEbiography}

\begin{IEEEbiography}[{\includegraphics[width=1in,height=1.25in,clip,keepaspectratio]{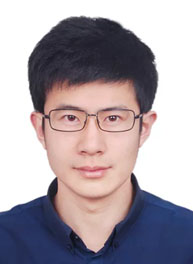}}]{Peng Wang} is a senior research scientist in Baidu USA LLC. He obtained his Ph.D. degree in University of California, Los Angeles, advised by Prof. Alan Yuille. Before that, he received his B.S. and M.S. from Peking University, China. His research interest is image parsing and 3D understanding, and vision based autonomous driving system. He has around 30 published papers in ECCV/CVPR/ICCV/NIPS.
\end{IEEEbiography}

\begin{IEEEbiography}[{\includegraphics[width=1in,height=1.25in,clip,keepaspectratio]{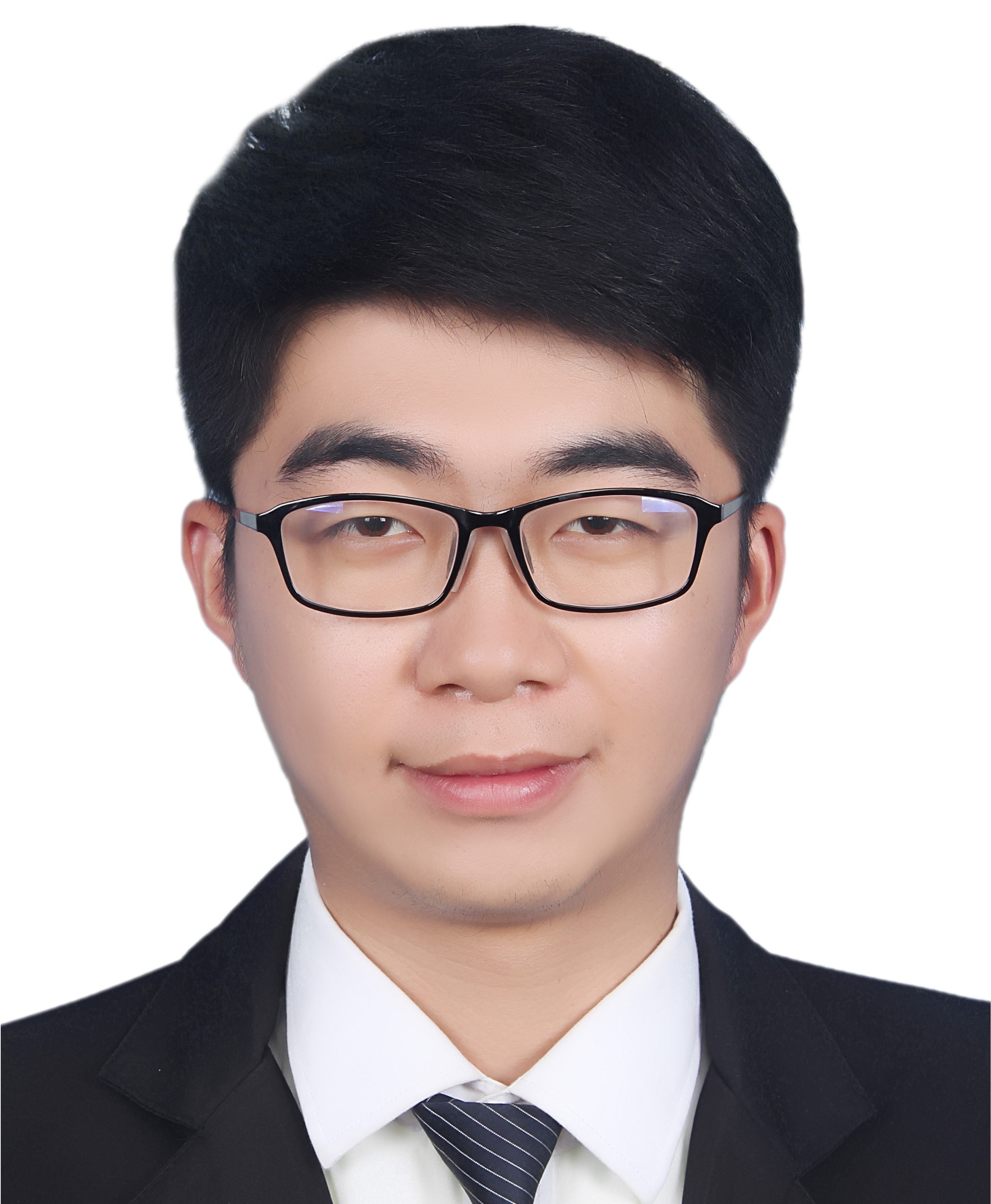}}]{Xinjing Cheng} is a research scientist with Robotics and Autonomous Driving Lab, Baidu Research, Baidu Inc., Beijing, China. Before that, he was a research assistant with the Intelligent Bionic Center, Shenzhen Institutes of Advanced Technology (SIAT), Chinese Academy of Sciences(CAS), Shenzhen, China. His current research interests include computer vision, deep learning, robotics and autonomous driving.
\end{IEEEbiography}

%\newpage

\begin{IEEEbiography}[{\includegraphics[width=1in,height=1.2in,clip,keepaspectratio]{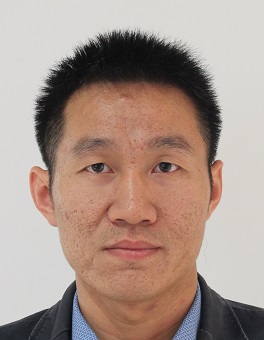}}]{Dingfu Zhou} a senior researcher at Robotics and Autonomous Driving Laboratory (RAL) of Baidu. Before joining in Baidu, he worked as a PostDoc Researcher in the Research School of Engineering at the Australian National University, Canberra, Australia. He obtained his Ph.D degree in System and Control from Sorbonne Universit\'{e}s, Universit\'{e} de Technologie de Compi\`{e}gne, Compi\`{e}gne, France, in 2014. He received the B.E. degree and M.E degree both in signal and information processing from Northwestern Polytechnical University, Xian, China. His research interests include Simultaneous Localization and Mapping, Structure from Motion, Classification and their application in Autonomous Driving.
\end{IEEEbiography}

\begin{IEEEbiography}[{\includegraphics[width=1in,height=1.2in,clip,keepaspectratio]{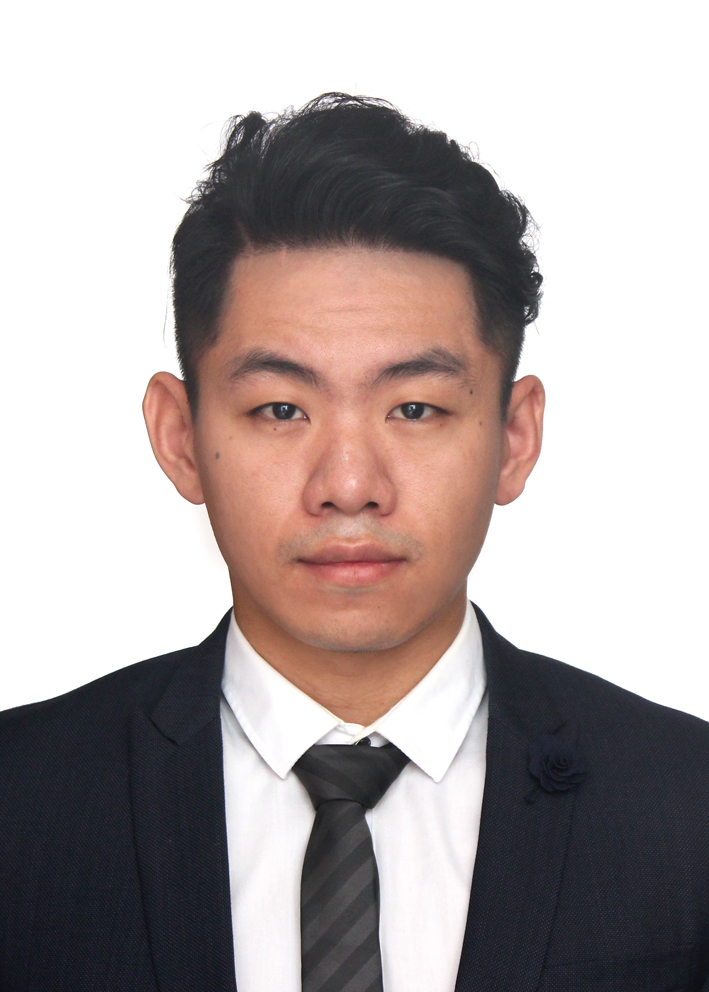}}]{Qichuan Geng}
  is a Ph.D. candidate, at State Key Lab of Virtual Reality Technology and Systems, Beihang University, Beijing, China. He
received his B.S. degree from Beihang University in 2012. His main research interests include computer vision, semantic segmentation and
scene geometry recovery.
\end{IEEEbiography}

\begin{IEEEbiography}[{\includegraphics[width=1in,height=1.2in,clip,keepaspectratio]{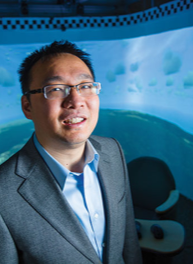}}]{Ruigang Yang} is the chief scientist for 3D vision at Baidu. He is also a full professor at the University of Kentucky (on leave). His research interests include 3D computer vision and 3D computer graphics, in particular 3D modeling and 3D data analysis. He has published over 100 papers with an H-index of 48. He is an Associate Editor for IEEE T-PAMI. He has been a program co-chair for 3DIMPVT (now 3DV) 2011 and WACV 2014, and he has been area chairs for both ICCV and CVPR multiple times.
\end{IEEEbiography}


%\vfill


\end{document}